\documentclass{article}

\usepackage{microtype}
\usepackage{graphicx}
\usepackage{subfigure}
\usepackage{booktabs} % for professional tables

\usepackage{hyperref}
\usepackage{float}

\usepackage[accepted]{icml2024}

\usepackage{amsmath}
\usepackage{amssymb}
\usepackage{mathtools}
\usepackage{amsthm}

\usepackage{csquotes}

\usepackage[capitalize,noabbrev]{cleveref}

\theoremstyle{plain}

\theoremstyle{definition}

\theoremstyle{remark}

\usepackage[textsize=tiny]{todonotes}

\usepackage{xspace}
\newcommand{\ph}{$\phi$\xspace}
\newcommand{\ps}{$\psi$\xspace}
\newcommand{\omone}{$\omega_1$\xspace}
\newcommand{\omtwo}{$\omega_2$\xspace}
\newcommand{\degree}{$^\circ$\xspace}

\icmltitlerunning{Enhanced sampling with uncertainty-based collective variable}

\begin{document}

\twocolumn[
\icmltitle{Enhanced sampling of robust molecular datasets with uncertainty-based collective variables}
% (four vote) Let Uncertainty Guide You: Diverse Data Sets for Robust Neural Network Interatomic Potentials
% (Rafa's vote) Uncertainty-Guided Enhanced Sampling for Robust Interatomic Potentials
% Optimizing Data Sets via Uncertainty-Guided Enhanced Sampling
% Reliable/Robust Interatomic Potentials from Uncertainty-Guided Sampling 
%(What about this?) Let uncertainty guide you: Automatic sampling of molecular free energy surfaces with uncertainty-based collective variables
% (This?) Let uncertainty guide you: Free energy surface sampling with uncertainty-based collective variable
% (Rafa's) Enhanced sampling of robust molecular datasets with uncertainty-based collective variables



\icmlsetsymbol{equal}{*}

\begin{icmlauthorlist}
\icmlauthor{Aik Rui Tan}{dmse}
\icmlauthor{Johannes C. B. Dietschreit}{dmse,vienna}
\icmlauthor{Rafael G\'omez-Bombarelli}{dmse}
\end{icmlauthorlist}

\icmlaffiliation{dmse}{Department of Materials Science and Engineering, Massachusetts Institute of Technology, Cambridge, MA, United States of America}
\icmlaffiliation{vienna}{University of Vienna, Vienna, Austria}

\icmlcorrespondingauthor{Rafael G\'omez-Bombarelli}{rafagb@mit.edu}

\icmlkeywords{Machine Learning, ICML}

\vskip 0.3in
]



\printAffiliationsAndNotice{}

\begin{abstract}
Generating a data set that is representative of the accessible configuration space of a molecular system is crucial for the robustness of machine learned interatomic potentials (MLIP). 
However, the complexity of molecular systems, characterized by intricate potential energy surfaces (PESs) with numerous local minima and energy barriers, presents a significant challenge. 
Traditional methods of data generation, such as random sampling or exhaustive exploration, are either intractable or may not capture rare, but highly informative configurations. 
In this study, we propose a method that leverages uncertainty as the collective variable (CV) to guide the acquisition of chemically-relevant data points, focusing on regions of the configuration space where ML model predictions are most uncertain. 
This approach employs a Gaussian Mixture Model-based uncertainty metric from a single model as the CV for biased molecular dynamics simulations. 
The effectiveness of our approach in overcoming energy barriers and exploring unseen energy minima, thereby enhancing the data set in an active learning framework, is demonstrated on the alanine dipeptide benchmark system. 
\end{abstract}

%%%%%%%%%%%%%%%%%%%%
%%% INTRODUCTION %%%
%%%%%%%%%%%%%%%%%%%%
\section{Introduction}
\label{introduction}
Computer simulations at the atomic scale are central for various scientific disciplines for elucidating the mechanisms that underpin key physical interactions in molecular and material systems\cite{graf_structure_2007,boese_ab_2003,kresse_efficient_1996,evans_computer_1983}.
Traditionally, two major families of methodologies have been used: i) \textit{ab initio} molecular dynamics, where the atomic forces are derived from quantum mechanical (QM) calculations, which tend to be highly accurate, but are computationally very expensive\cite{kresse_ultrasoft_1999,perdew_density-functional_1986} and ii) empirical force fields simulations, which offer computational efficiency at the expense of accuracy\cite{urata_molecular_2021,beest_force_1990,tsuneyuki_first-principles_1988}, as they are based on classical expressions fitted to, e.g., QM data. 
To bridge the gap between accuracy and computational feasibility, machine learned interatomic potentials (MLIPs) have emerged as a powerful tool to enhance the predictive modeling of complex interactions\cite{keith_combining_2021,deringer_machine_2019,butler_machine_2018,schwalbe-koda_generative_2020,mueller_machine_2020}. 
MLIPs have been extensively applied in molecular simulations of proteins\cite{wang_differentiable_2020}, small molecules\cite{jose_construction_2012}, water\cite{schran_transferability_2021}, solids\cite{calegari_andrade_structure_2020}, and more\cite{podryabinkin_active_2017}. 

However, MLIPs essentially function as advanced interpolation schemes, relying on the quality and breadth of the training data from which they learn. 
This dependence presents a great challenge regarding robustness and generalizability of the MLIPs, particularly within the extrapolative regimes where the training data is sparse\cite{xu_how_2020,fu_forces_2022,vita_data_2023}. 
Consequently, for MLIPs to facilitate reliable and insightful dynamics, the span of the training dataset across configuration space is critical. 
The dataset should ideally encompass a diverse and comprehensive range of configurations to ensure the robust application of MLIPs in predictive modeling. 
Nevertheless, building suitable training data poses a significant challenge. 
Many data sets such as rMD17 are generated via classical molecular dynamics (MD) simulations with QM calculations performed on selected trajectory frames. 
Such data sets tend to sample configurations based on the thermodynamic probability given by the potential energy surface (PES) and focus mostly on configurations near the energy minima. 
This approach neglects configurations important for rare events, leading to simulations that are unphysically trapped in minima, thereby generating unreliable statistics and creating nonviable structures, as the model has no information about their inherent high energy\cite{morrow_how_2023}. 

An alternative method to enrich training data involves the integration of active learning with uncertainty quantification, steering data acquisition towards regions of high epistemic uncertainty, which are usually a byproduct of limited data\cite{schwalbe-koda_differentiable_2021}. 
Such a strategy is pivotal for broadening the scope of data exploration into areas where MLIP predictions are the least reliable. 
By iteratively expanding the training set towards high uncertainty regimes, active learning aims to explore configuration space well while at the same time reducing the extrapolative errors\cite{ang_active_2021,shapeev_active_2020}. 

In this work, we propose to generate a diverse training set by combining existing enhanced sampling methods with the MLIP uncertainty estimate as collective variable (CV, or reaction coordinate). 
This allows us to bias the system towards key areas which are undersampled in the current training data, characterized by higher uncertainty values, and thus to improve MLIP robustness. 
We employ the extended-system adaptive biasing force (eABF)\cite{chen_overcoming_2021} coupled with Gaussian-accelerated molecular dynamics (GaMD)\cite{miao_gaussian_2015} methods for enhanced sampling. 
One key advantage of using uncertainty as the CV is eliminating the need to pre-define a reaction coordinate specifically tailored to capture the slow degrees of freedom for a system. 
We posit that the estimated uncertainty serves as a versatile CV, effectively encompassing all degrees of freedom. 
On the other hand, similar approaches\cite{kulichenko_uncertainty-driven_2023,van_der_oord_hyperactive_2023,zaverkin_uncertainty-biased_2023} have previously employed predicted uncertainties from MLIPs, they predominantly utilized uncertainty as the biasing energy rather than as a CV. 
These strategies require the predicted uncertainty to align in both unit and magnitude with the potential energy of the system, thereby imposing constraints on the applicable uncertainty quantification techniques. 
In addition, our approach balances the dual objective of exploration and exploitation without generating configurations that are too distorted or too similar due to mode collapse, a drawback observed in a previous methodology\cite{schwalbe-koda_differentiable_2021,tan_single-model_2023}. 
Furthermore, we use single model uncertainty as opposed to uncertainty derived from an ensemble, thereby reducing the training and exploration cost. 
We demonstrate the efficacy of our approach on creating a data set for alanine dipeptide, a typical benchmark molecule for sampling techniques due to its complex intramolecular motions\cite{mironov_systematic_2019,wei_conformational_2001}.

To summarize, our contributions include:

\begin{itemize}
    \item An approach that harnesses single-model uncertainty as the collective variable in enhanced sampling methods. This approach is designed to create a diverse data set with good coverage of configuration space. 
    \item Enhanced efficacy of our method as demonstrated on the flexible alanine dipeptide molecule using minimal initial training data in the active learning setting. 
\end{itemize}

% CRITICISM:  Flesh out the theoretical basis for this choice more thoroughly. Discussing the theoretical implications and potential limitations of the approach can add depth to the study.

%%%%%%%%%%%%%%%%%%%%
%%% RELATED WORK %%%
%%%%%%%%%%%%%%%%%%%%
\section{Related Work}

\label{sec:RelatedWork}

\textbf{Uncertainty quantification}. In the field of computational materials science and chemistry, uncertainty quantification is widely used to enhance model reliability\cite{peterson_addressing_2017}, guide experimental design\cite{honarmandi_uncertainty_2020}, and identify key factors of variability\cite{oreluk_diagnostics_2018}. 
Since neural networks (NNs) do not intrinsically provide uncertainty estimates for their predictions, various strategies such as Bayesian NNs\cite{thaler_scalable_2022,vandermause_--fly_2020}, Monte Carlo dropout\cite{gal_dropout_2016}, NN ensembles\cite{lakshminarayanan_simple_2017}, and single-model uncertainty\cite{janet_quantitative_2019,tan_single-model_2023} methods are employed so that the NNs predictions are `confidence-aware'. 
Bayesian NNs, while capable of providing uncertainty estimates from the distribution of model parameters, can be prohibitively expensive for use as MLIPs for complex systems. 
Similarly, NN ensembles, often considered the golden standard, present a computational burden due to the necessity of training and inferring from multiple independent NNs at run time. 
Hence, recent research has increasingly focused on single deterministic model uncertainty methods. 
These approaches enable the evaluation of uncertainty through a single forward pass of the data, thereby eliminating the need for multiple model usage or stochastic sampling to approximate the underlying uncertainty functions. 
One of these methods include the application of Gaussian mixture model (GMM) to the latent space of a single NN, and deriving uncertainty estimates based on the \enquote{distance} of new data points from the centers of the Gaussians. This method not only proves to be a reliable uncertainty quantification method, but also remains computationally efficient\cite{zhu_fast_2023}.

\textbf{Uncertainty-maximized active learning}. The concept of harnessing uncertainty as a catalyst for exploring the configuration space has gained significant traction in recent years. 
It operates on the premise that areas of high uncertainty often harbor key insights or novel phenomena that are crucial for a deeper understanding of complex systems. 
This approach has been effectively demonstrated in a variety of contexts, showcasing its versatility and impact.

\citeauthor{schwalbe-koda_differentiable_2021} proposed an adversarial sampling method combining adversarial attacks and differentiable uncertainty metrics to sample high thermodynamic likelihood and high uncertainty configurations, thereby eliminating the need for MD to generate new geometries. Instead of performing gradient descent, the methodology employs gradient ascent on estimated uncertainty by incrementally distorting the input geometries. In other applications, as detailed in the papers by \citeauthor{kulichenko_uncertainty-driven_2023} and \citeauthor{van_der_oord_hyperactive_2023}, a MD sampling methodology was proposed, where the PES is biased in direct correlation with the estimated uncertainty of the configuration. These strategies predominantly derive uncertainty estimates from an ensemble of machine learning (ML) models. Complementarily, \citeauthor{zaverkin_uncertainty-biased_2023} employed a similar concept of uncertainty-based biasing energy to expand the dataset, albeit relying on uncertainty estimates from a singular model and applies a biasing stress on top of a biasing potential.

%%%%%%%%%%%%%%%%%%%
%%%%% METHOD %%%%%%
%%%%%%%%%%%%%%%%%%%
\section{Uncertainty-guided enhanced sampling}

\subsection{Neural network interatomic potentials (NNIPs)}
This work uses neural network interatomic potentials (NNIPs) to fit the PES, but our method can be generalized to other MLIP architectures that possess a diferentiable uncertainty metric.
All NNIPs were trained using the MACE architecture, which at its core has an equivariant message-passing scheme that incorporates higher body order messages\cite{batatia_mace_2023}.
In this work, the MACE models used to run uncertainty-guided enhanced sampling are made up of 4-channel dimension layers, whereas for production runs and generation of potential of mean force (PMF), the layers are made of 16-channel dimension. Such small model for uncertainty-guided enhanced sampling is used to accelerate exploration of phase space. For more details on the hyperparameters used, refer to Section \ref{sec:mace}.

%%%%%%%
% GMM %
%%%%%%%
\subsection{Gaussian mixture model (GMM) based uncertainty} 
A GMM represents the data set, $x$ as subpopulations characterized by a weighted mixture of $K$ Gaussian distributions\cite{reynolds_gaussian_2009}, as described below
\begin{equation}
    p(x|\pmb{w}, \pmb{\mu}, \pmb{\Sigma}) = \sum_{k=1}^K w_k \mathcal{N}(\pmb{\mu}_k, \pmb{\Sigma}_k).
\end{equation}
Each single Gaussian distribution $k$ of dimension $D$ is parameterized by its mean vector, $\mu_k \in \mathbb{R}^{D\times 1}$ and its covariance matrix, $\Sigma_k\in\mathbb{R}^{D\times D}$. 
The sum of all non-negative mixture weights, $w_k$, is 1. 
To estimate the uncertainty of a molecular configuration, we fit a GMM to the distribution of latent atomic features formed by all the configurations in the training set, $\zeta_{\text{train}}\in \mathbb{R}^{D\times 1}$, using the expectation-maximization (EM) algorithm with full-rank covariance matrices and mean vectors predetermined using k-means clustering\cite{zhu_fast_2023}. 
Once fitted, uncertainty of a trial data point is estimated as the negative log-likelihood $\text{NLL}(\zeta_{\text{test}}|\zeta_{\text{train}})$.
\begin{equation}
    \text{NLL}(\zeta_{\text{test}}|\zeta_{\text{train}}) = - \log \left( \sum_{k=1}^K w_k \mathcal{N}(\zeta_{\text{test}} | \mu_k, \Sigma_k) \right) \label{eq:gmm_unc}
\end{equation}
where $\zeta_{\text{test}}$ indicates the atomic latent features of the test data. 
In this work, the GMM is fitted to the atomic latent feature, thus the uncertainty for a molecular configuration is taken to be the mean of the per-atom NLL, $\text{NLL} = \frac{1}{N_\text{atoms}} \sum_i^{N_\text{atoms}} \text{NLL}_i$. 
Since $\text{NLL}(\zeta_{\text{test}}|\zeta_{\text{train}})$ measures how far removed the latent embedding of a configuration is from the mean vectors of the GMM, a higher $\text{NLL}(\zeta_{\text{test}}|\zeta_{\text{train}})$  corresponds to a higher uncertainty.

%%%%%%
% CP %
%%%%%%
\subsection{Conformal Prediction} \label{sec:cp}
Conformal prediction (CP) calibrates confidence level of new predictions by utilizing quantile regression on previous observations. The framework of CP applied to guarantee calibration and finite sample coverage can be defined as\cite{angelopoulos_gentle_2022}
\begin{equation}
    P\{ Y_\text{new} \in C(X_\text{new}) \} \geq 1 - \alpha
\end{equation}
where $(X_\text{new}, Y_\text{new})$ is the new data, $C$ is the prediction set based on previous observations, and $\alpha$ defines the desired confidence level. 
Since GMM-based uncertainty is a heuristic notion for the true error and is not intended to have the same unit as the error, the uncertainty value can be an over- or underestimate compared to the true error\cite{hirschfeld_uncertainty_2020}. 
Hence, CP is utilized to better calibrate the estimated uncertainty to the true error. 
In accordance to \citeauthor{hu_robust_2022}, we employed inductive CP by first randomly choosing $n=100$ data points to be our calibration set, and then calculating the GMM-uncertainty, $\mathbf{u} \in \mathbb{R}^{n\times 1}$ as well as the true error, $\pmb{\varepsilon} \in \mathbb{R}^{n\times 1}$ from the NNIP's predictions as $\frac{1}{N_\text{atoms}} \sum_i^{N_\text{atoms}} ||\mathbf{F}_{i} - \mathbf{\hat{F}}_{i}||_2$. Using quantile regression, the $n$ samples are used to obtain a score function, $\mathbf{s} = \frac{\pmb{\varepsilon}}{\mathbf{u}}$, which yields the scaling factor $\hat{q}=\frac{(n+1)(1-\alpha)}{n}$. 
The calibrated uncertainty, $\tilde{\mathbf{u}}$ is then computed as $\tilde{\mathbf{u}} = \hat{q} \times \mathbf{u}$, where $\hat{q}$ is simply a constant. We chose an alpha, $\alpha$ value of 0.05 for all experiments (See Figure~\ref{fig:cp_alpha} for CP experiments).

%%%%%%%%%%%%%
% eABF-GaMD %
%%%%%%%%%%%%%
\subsection{Uncertainty as CV in eABF-GaMD}

The extended-system adaptive biasing force (eABF) method is a common technique used in MD simulations to overcome barriers in free energy landscapes. It improves upon the original ABF approach\cite{comer_adaptive_2015} by using a fictitious variable, $\lambda$ that is harmonically coupled to the collective coordinates of interest, rather than directly biasing these coordinates\cite{fu_extended_2016}, which allows for more flexible manipulation of the system. 
In principle, any history dependent biasing technique that pushes the system to higher uncertainty regions but also allows it to return to known regions could be used here. 
For a backdrop on the details of eABF, see Section~\ref{sec:eabf-gamd}.
The extended potential, $V_\text{ext}$ applied to the system is
\begin{equation}
    V_{\text{ext}}(\mathbf{x}, \lambda, t) = V(\mathbf{x}) + \frac{1}{2\beta\sigma^2}(\xi(\mathbf{x}) - \lambda)^2 + V_\text{ABF}(\lambda, t)
\end{equation}
where $V(\mathbf{x})$ denotes the PES, $V_\text{ABF}(\lambda, t)$ is the potential afforded by ABF, $\beta = 1/k_BT$, $\sigma$ is the \enquote{thermal coupling width} between the extended system and the CV of interest, $\xi(\mathbf{x})$ for the configuration $\mathbf{x}$.
Here, we seek to replace the CV of interest with the predicted
uncertainty, $\xi(\mathbf{x}) = u(\mathbf{x})$. 
A small $\sigma$ value corresponds to a tight coupling between the CV (here the uncertainty) and the fictitious particle. 
This approach avoids the need for second derivative estimates, making eABF more broadly applicable. It also provides a smoother, coarse-grain-like sampling that allows for faster exploration\cite{comer_adaptive_2015}. 
To further improve the sampling efficiency, we follow the approach from \citeauthor{chen_overcoming_2021} by combining eABF with Gaussian-accelerated MD (GaMD), which adds a harmonic boost potential without requiring predefined CVs, allowing for faster escape from local minima (details can also be found in Sec.~\ref{sec:eabf-gamd}). 

In all uncertainty-guided enhanced sampling simulations, the initial atomic configurations for all simulations were chosen randomly from the training sets of each generation. Relaxation of atomic positions of the initial atomic configurations were performed for 10 steps using the FIRE algorithm\cite{bitzek_structural_2006}. In each active learning iteration, two uncertainty-guided eABF-GaMD simulations were performed, one at 300~K and the other 500~K. In the first active learning iteration, the simulations are run for 50,000 steps (12.5~ps) with a thermal coupling constant of $\sigma_{i=1} = 0.5$. 
With each subsequent iteration, the total number of steps is increased by a factor of 1.2 whereas $\sigma_{i}$ is decreased by a factor of 0.8 ($\sigma_{i} = \sigma_{i=1} \times 0.8^{i-1}$), tightening the coupling of uncertainty to the fictitious particle. See Section \ref{sec:eabf-gamd} for more parameter specifications and changes during the active learning iterations.

%%%%%%%%%%%%%%%
% Acquisition %
%%%%%%%%%%%%%%%
\subsection{Data acquisition}
After conducting simulations at 300~K and 500~K in each active learning iteration, we selected configurations that exhibited predicted uncertainties exceeding a defined threshold, denoted as $u_\text{cutoff}$. Subsequently, a handful of representative configurations were identified by applying hierarchical clustering to the cosine distances of their latent features, as derived from the MACE models. This process ensured diversity, as only one representative configuration was randomly chosen from each cluster, effectively preventing repetitive sampling and redundant ground truth evaluations of similar geometries. In the first active learning iteration, we set $u_\text{cutoff} = 2.0$ (approximately equivalent to 2.0~eV post inductive CP), and incrementally increased it by a factor of 1.05 after every iteration. For the hierarchical clustering, we employed the maximum (complete) linkage method with a minimum cosine distance threshold of 0.015.

%%%%%%%%%%%%%%%%%%%
%%% EXPERIMENTS %%%
%%%%%%%%%%%%%%%%%%%

\begin{figure}[!ht]
    \centering
    \includegraphics[width=\linewidth]{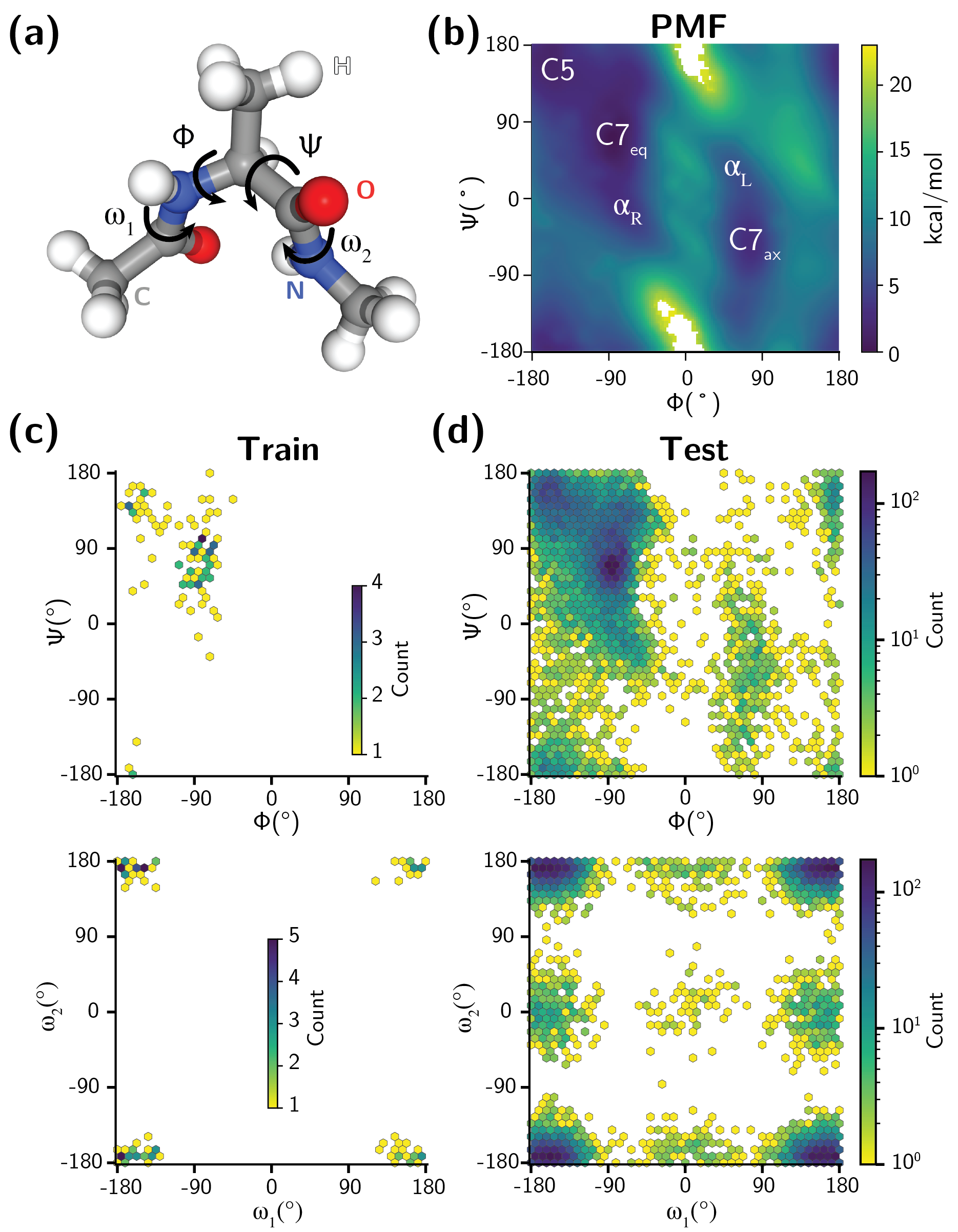}
    \vspace{-5ex}
    \caption{%
    \textbf{(a)},  Structure of the alanine dipeptide molecule with carbon (C), nitrogen (N), oxygen (O), and hydrogen (H) atoms labeled in grey, blue, red, and white, respectively. Four backbone dihedral angles \ph, \ps, \omone, and \omtwo are annotated.
    \textbf{(b)}, Potential mean force (PMF) profile of the N-acetyl-L-alanine-N methylamide (alanine dipeptide) molecule on the backbone dihedral angles \ph and \ps, provided by the umbrella sampling method using the amber ff19SB force field\cite{tian_ff19sb_2020}. Regions with stable conformations are labeled\cite{mironov_systematic_2019}.
    \textbf{(c)}, Top subfigure shows Ramachandran plot of 100 configurations provided as the initial data set to train NNs in the first generation. Bottom subfigure shows distribution of the same data set but plotted with respect to the \omone and \omtwo backbone dihedral angles.
    \textbf{(d)}, Top and bottom subfigures show distributions of test data set on the \ph-\ps and \omone-\omtwo backbone dihedral angles, respectively. NNs from all generations are not trained or validated on any data from the test set. 
    }
    \label{fig:fig1}
\end{figure}

\section{Results}

Here, we demonstrate the  acquisition of diverse data afforded by the uncertainty-guided eABF-GaMD technique on the N-acetyl-L-alanine-N methylamide (or more commonly known as alanine dipeptide). 
%In this work, the isomerization of the alanine dipeptide molecule in vacuum is explored. 
Alanine dipeptide (Fig \ref{fig:fig1}a), with its apparent simple structure, is a system frequently used for enhanced sampling benchmarks due its complex intramolecular motions. 
Conformational changes of the molecule are typically studied with respect to the \ph (C-N-C$_\alpha$-C) and \ps (N-C$_\alpha$-C-N) backbone dihedral angles (as denoted in Fig \ref{fig:fig1}a), but the peptide bond dihedrals \omone (C$_\alpha$-N-C(=O)-C$_\omega$) and \omtwo (C$_\alpha$-C(=O)-N-C$_\omega$), especially \omone, have also been found to be important for conformational analysis\cite{bolhuis_reaction_2000}. 

As reference, we present the ground truth PMF of alanine dipeptide in the \ph-\ps plane, based on the Amber ff19SB force field\cite{tian_ff19sb_2020}, in Fig.~\ref{fig:fig1}b (See Section \ref{sec:umbrella} for PMF generation details). 
A well trained NNIP is expected not only to run stable MD simulations and have a low testing error, but also able to reproduce this PMF.
Fig.~\ref{fig:fig1}b clearly displays several distinct energy basins, which correspond to ensembles of stable configurations, most notably C7$_{eq}$ (\ph$\approx$-90\degree,\ps$\approx$70\degree), C5 (\ph$\approx$-160\degree,\ps$\approx$160\degree), C7$_{ax}$ (\ph$\approx$70\degree,\ps$\approx$-60\degree), $\alpha_R$ (\ph$\approx$-90\degree,\ps$\approx$-10\degree), and $\alpha_L$ (\ph$\approx$60\degree,\ps$\approx$30\degree)\cite{chen_heating_2012}.
These minima are also frequently visited by the unbiased NVT simulations at elevated temperatures (500~K and 1200~K), as illustrated in the upper panel of Fig.~\ref{fig:fig1}d. 

The bottom panels of Figs.~\ref{fig:fig1}c and \ref{fig:fig1}d, show the distributions of configurations in the \omone-\omtwo backbone dihedral angles. 
Especially at lower temperatures, the peptide bond angles \omone and \omtwo are both close to 180\degree, whereas the \omone=0\degree or \omtwo=0\degree states become accessible only at higher temperatures, signifying the strong preference for planar configurations for these bonds. 
This behavior is attributed to the delocalization of the lone electron pair on the nitrogen atom into the carbonyl group, resulting in a partial double bond that increases the rotational barrier of the peptide bond\cite{mironov_systematic_2019}. 

For the initial training and validation set, we randomly selected only 100 configurations each from the cooler 500~K simulation (Fig.~\ref{fig:fig1}c), whereas the test set (Fig.~\ref{fig:fig1}d) contains all remaining 9800 configurations from both 500~K and 1200~K simulations (see Section \ref{sec:datasets} for more details).
It is noteworthy that for the training set, the configurations are mostly contained within the C7$_{eq}$ basin and with \omone=-180\degree or 180\degree and \omtwo=-180\degree or 180\degree. 
This localization underscores the exploratory power of our method in expanding beyond these initial bounds.

\subsection{Efficacy of uncertainty-guided enhanced sampling}

\begin{figure}[t]
    \centering
    \includegraphics[width=\linewidth]{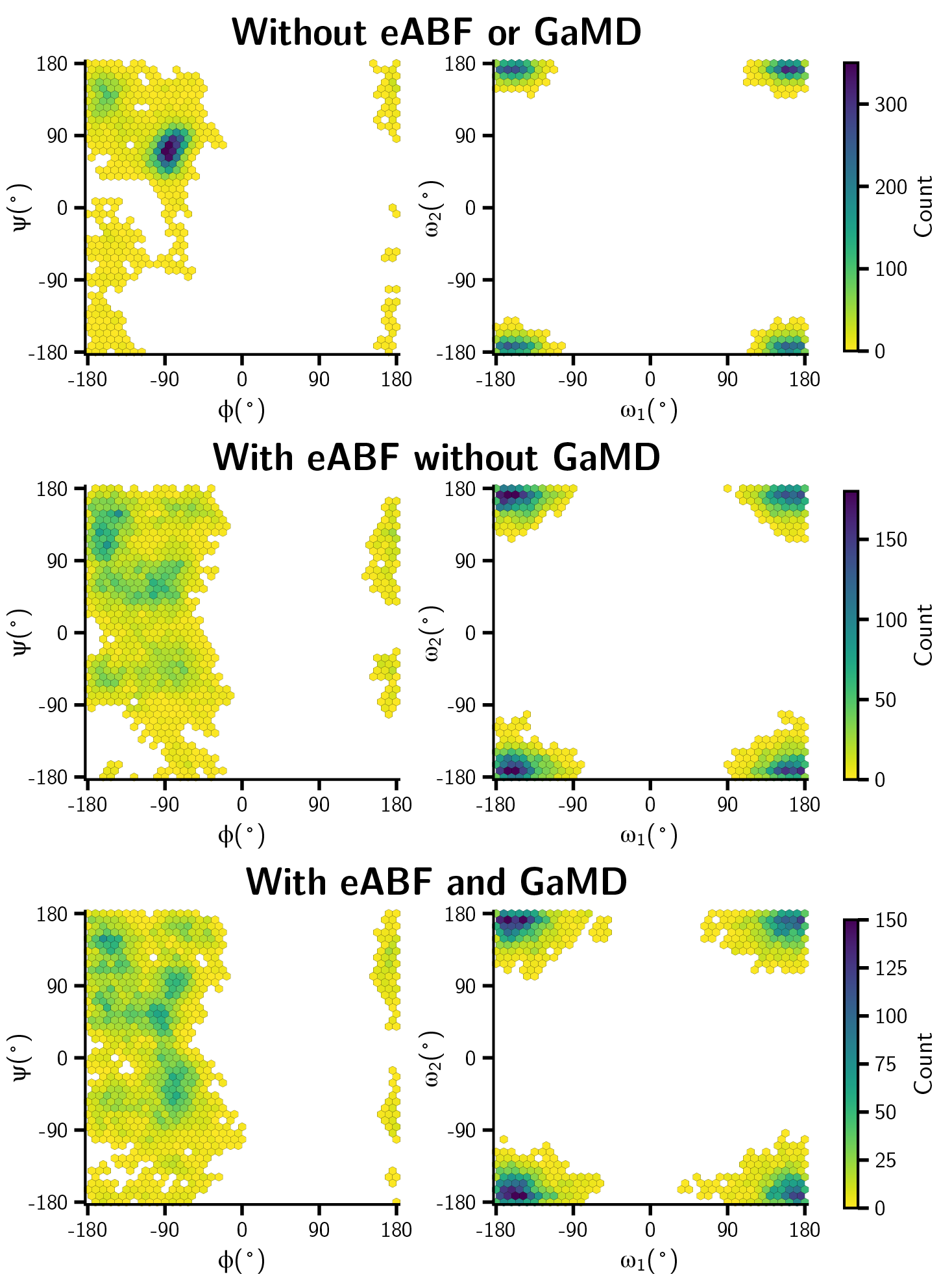}
    \vspace{-5ex}
    \caption{Cumulative exploration of configuration space projected onto the \ph-\ps (left column) and \omone-\omtwo (right column) plane of 10 NVT simulations at 300~K. 
    Top) No biasing of any sort,  
    middle) uncertainty-guided eABF, and bottom) uncertainty-guided eABF-GaMD.}
    \label{fig:efficacy_eabf-gamd_300K}
\end{figure}

To assess the efficacy of the uncertainty-guided eABF-GaMD approach, we compare the exploration of the \ph-\ps space during simulations with distinct setups: unbiased NVT, biased with eABF, and biased with eABF-GaMD (as illustrated in Fig.~\ref{fig:efficacy_eabf-gamd_300K}). 
We conducted ten simulations for each scenario at 300~K to ensure that the coverage of the CV space was not due to random fluctuations, but rather indicative of a consistent pattern. 
These simulations were executed using the smaller, 4-channel MACE model trained exclusively on the initial 100 point data set depicted in Fig.~\ref{fig:fig1}c.

The top row of Fig.~\ref{fig:efficacy_eabf-gamd_300K}, where the simulations were unbiased, shows a predominant confinement to the C7$_{eq}$ basin. 
However, with the incorporation of uncertainty-guided eABF and eABF-GaMD, a substantial increase in exploration was observed, successfully discovering regions like C5 and $\alpha_R$ (\ph$\approx$-60\degree,\ps$\approx$45\degree) basins.
The addition of GaMD has also increase the explored regions considerably.
Remarkably, the uncertainty CV also allowed for the slight rotation of the more rigid \omone and \omtwo dihedral angles towards 90\degree/-90\degree at just 300~K, without resulting in unphysical geometries. 
This underscores the efficacy of the uncertainty CV in directing molecular exploration towards extrapolative regions, while concurrently maintaining physicality of the configurations. 
Similar comparison of the explored phase space with setups of unbiased, eABF, and eABF-GaMD simulations at 500~K is shown in Fig \ref{fig:comparison_eabf-GaMD-500K}, where one can observe a stronger stratifying effect of GaMD.

%%%%%%%%%%%%%%%%%%%%%%%%%%%%%%%%%%%%%%%%%%%%%%

\begin{figure*}[t]
    \centering
    \includegraphics[width=\linewidth]{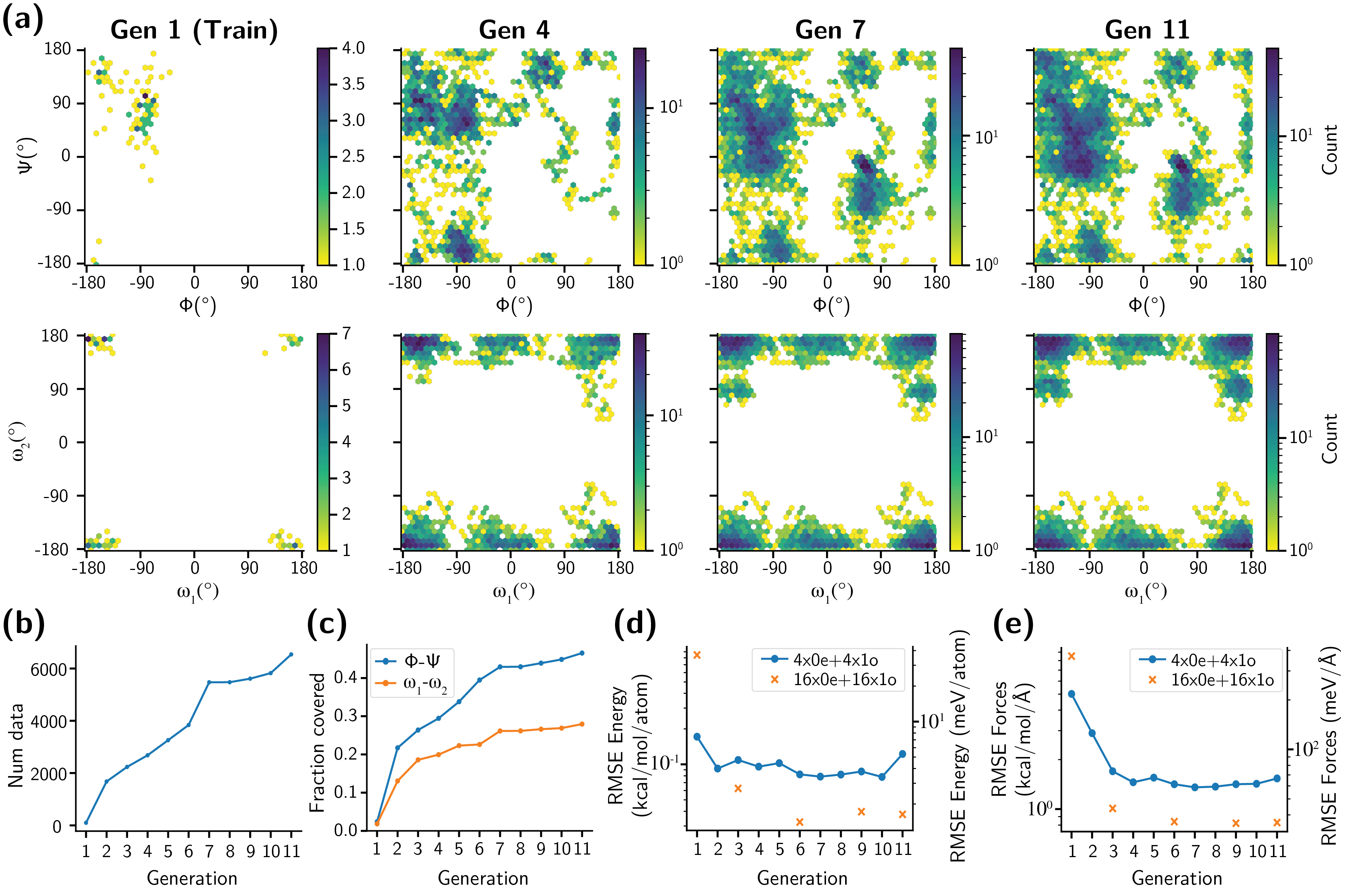}
    \vspace{-5ex}
    \caption{%
    \textbf{(a)}, Hexbin plots representing data sets accumulative over generations of active learning, showcasing the distribution of \ph-\ps (top) and \omone-\omtwo (bottom) backbone dihedral angles during GMM-based uncertainty-guided eABF-GaMD simulations. 
    \textbf{(b)}, Number of total data points used for training the NNIPs in each generation. The initial value of 100 points at generation 1 indicates number of initial training data shown in Fig \ref{fig:fig1}a.
    \textbf{(c)}, Fraction of grid-based coverage of the \ph-\ps and \omone-\omtwo space in each generation. Horizontal dashed lines describes coverage for the test set. 
    \textbf{(d)} and \textbf{(e)}, Mean absolute error of predicted energy and forces of configurations in the test set as generations proceed. Blue points indicate predictions from NNIPs of smaller MACE models with 4-channels, whereas orange points indicate predictions from the bigger models with 16-channels. The two axes in each plot correspond to the most common units used for atomistic simulations with MLIPs. 
    }
    \label{fig:perf}
\end{figure*}

\subsection{Active learning}

Fig.~\ref{fig:perf}a shows the cumulative sampled configurations throughout the iterations in one active learning experiment.
% All active learning experiments have been repeated 5 times. 
An active learning experiment in this case involves a sequence of steps starting with training NNIPs. This is followed by conducting uncertainty-guided eABF-GaMD simulations, collecting new data points from these simulations, carrying out ground truth calculations on the newly gathered data, and finally retraining the NNIPs with the updated dataset.
From iteration~1 (Gen~1) to iteration~4 (Gen~4), the explored configuration space increased the most, covering a wide region of the C7$_{eq}$ and C5 basins, and extended exploration to the \ph $>$ 0\degree regions. This underscores the biasing and exploration efficacy of the uncertainty-guided eABF-GaMD simulations within such short simulation times (e.g. 12.5~ps in Gen~1).
Furthermore, the uncertainty bias managed to rotate the \omone peptide bond to 0\degree. 
From Gen~4 to Gen~7, the C7$_{ax}$ basin was discovered, and there is an increased coverage in all the known energy basins and also into the non-planar \omtwo dihedral angles. 
Beyond Gen 7, the \ph-\ps coverage somewhat saturated. 
Most configurations with PMF values above 12.5~kcal/mol are not explored during the short runs in each iteration and there remains also a lack of coverage in the \omtwo = 0\degree dihedral region. 
% The lack of exploration in these region can have multiple reasons.
% First, the simulation time is simply too short for a strong enough bias to build up and explore these high energy regions (the inversion of \omtwo has a much larger barrier than \omone).
The lack of exploration could be attributed to the fact that 
the simulations are aborted whenever configurations with atomic distances smaller than 0.75~{\AA} are encountered, as such unphysical configurations have high uncertainty.

\begin{figure*}[t]
    \centering
    \includegraphics[width=\linewidth]{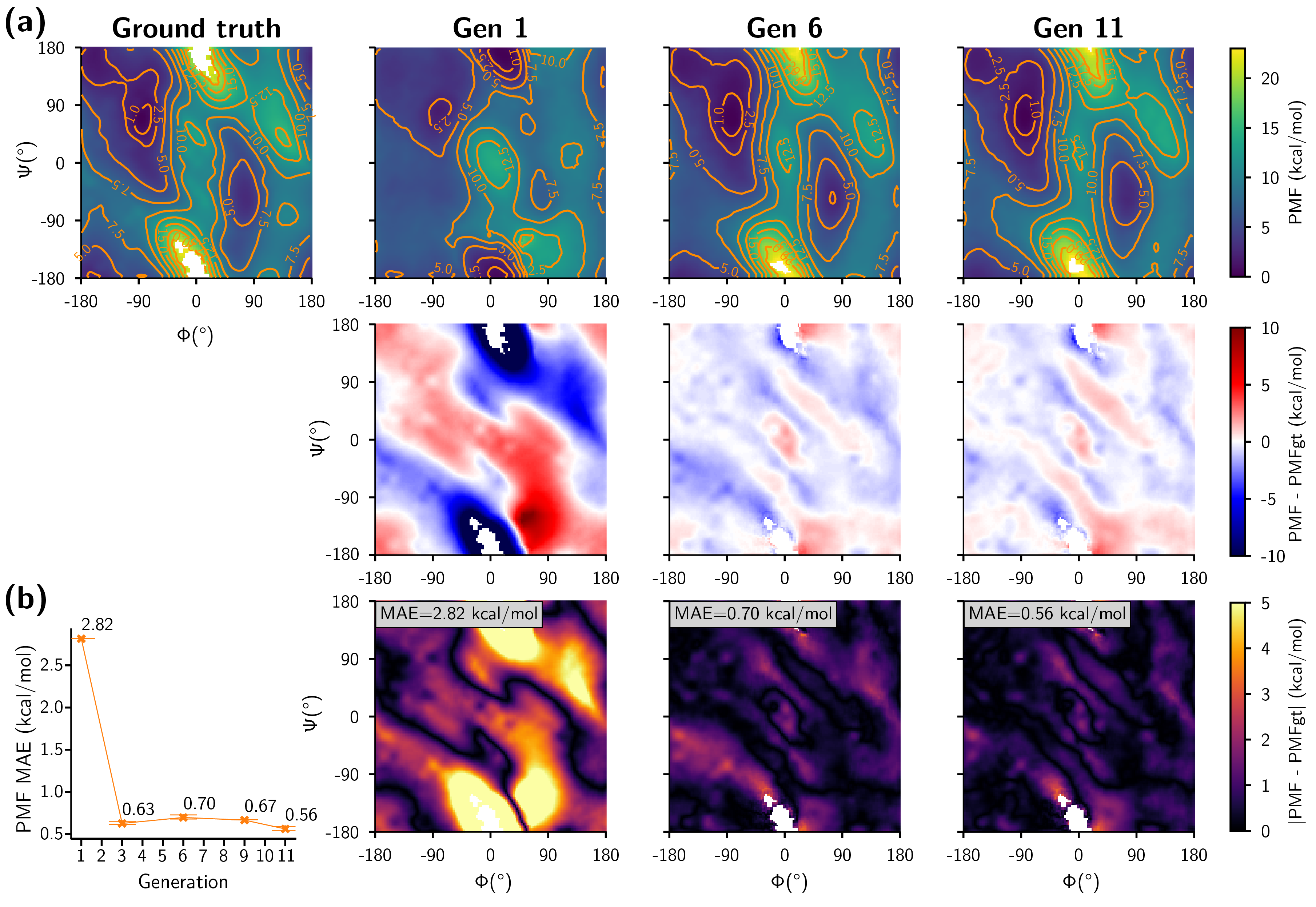}
    \vspace{-5ex}
    \caption{%
    \textbf{(a)}, Plots in the top row show the PMF profiles generated from the Amber ff19SB force field (Ground truth) and NNIPs in generation 1, 6, and 11, in the \ph-\ps backbone dihedral angles. Middle row plots show the differences of NNIPs-generated PMF contours from the ground truth PMF. Red and blue regions indicate PMF overestimation and underestimation, respectively. Plots in the bottom row show the absolute value of the PMF disparities (middle row), and the values shown on the top left of the plots indicate the mean absolute error (MAE) of the NNIPs-generated PMF profiles. Note that all NNIPs used to generate the PMF contours are based on 16-channel MACE models.
    \textbf{(b)}, Mean absolute errors (MAE) of the PMF profiles generated from NNIPs in generation 1, 3, 6, 9, and 11, with error bars showing ranges of MAEs.
    }
    \label{fig:pmfs}
\end{figure*}

To evaluate the effectiveness of our method in enhancing NNIP robustness, we analyzed its accuracy on the large test set. 
This set features extensive coverage in both \ph-\ps and \omone-\omtwo dihedral spaces (Fig.~\ref{fig:perf}c), resulting from the high temperature of 1200~K in the unbiased simulation. 
As seen from Fig.~\ref{fig:perf}d and \ref{fig:perf}e, the per atom energy error plateaus as early as Gen~2 for the small 4-channel NNIPs, showing only marginal improvements thereafter. 
The force error, on the other hand, stagnates after Gen~4, suggesting that the NNIPs, having been trained on the sampled data sets, can accurately predict energy and force values for most configurations in the test set. 

The larger 16-channel model exhibits a significantly higher error in Gen~1 compared to the 4-channel model, indicating a tendency towards over-fitting to the initial, limited training set. 
Intriguingly, the 16-channel model's error decreases further past Gen~3, even though the accuracy of the 4-channel model plateaus. 
This pattern suggests that the eABF-GaMD method, driven by uncertainty CV, continues to enhance sampling coverage past generation 4, even when the model used for sampling is very small and such improvements are not immediately apparent in the Ramachandran plots or the test set accuracy of small models past generation 4.

To ensure that there is indeed increase in enhanced sampling coverage, we calculate the grid-based coverage fraction for the \ph-\ps and \omone-\omtwo dihedral space (Fig.~\ref{fig:perf}c). Grid-based coverage considers a specified grid filled if a configuration with the corresponding dihedral CV exists within the dataset. Fraction of grid-based coverage can then be obtained by dividing the total number of filled grids points by their total number. The grid size here is 6\degree, which yields a total number of 3600 bins. As can be seen from Fig.~\ref{fig:perf}c, the coverage fraction for both \ph-\ps and \omone-\omtwo dihedral increases steeply from Gen~1 to Gen~3, followed by a lower rate of increase beyond Gen~3. There is a clear correlation between increased dihedral space coverage and improved predictive accuracy on the test set, especially for the bigger 16-channel model. It is also important to note that while the fraction of dihedral coverage is a useful metric, there are degrees of freedom in the diversity of new samples lost in the \ph-\ps and \omone-\omtwo projections. In total, the number of sampled data in this active learning experiment is less than 7000 data points (Fig.~\ref{fig:perf}b), which is significantly less than 38,000 points used for the same molecule in the paper by \citeauthor{fu_forces_2022}

In addition to test set accuracy, we performed umbrella sampling to generate PMFs from the NNIPs at different generations to show a more rigorous assessment. The PMFs for generations 1, 6, and 11 are depicted in the top row of plots in Fig.~\ref{fig:pmfs}, with the \enquote{ground truth} PMF serving as a reference in the first column. In generation 1, the NNIP accurately identified the C7$_{eq}$ minima, reflecting the influence of the training set configurations. Interestingly, the model also inferred the presence of an energy barrier at \ph$\approx$0\degree,\ps$\approx$0\degree, a small $\alpha_L$ energy basin near \ph$\approx$60\degree,\ps$\approx$30\degree, and the C7$_{ax}$ basin at \ph$\approx$70\degree,\ps$\approx$-60\degree, despite the absence of training data in all these regions. Nonetheless, the model inaccurately predicted an energy basin near \ph=0\degree,\ps=$\pm$180\degree, a region which should realistically be inaccessible due to steric hindrance. The model was also unable to predict the presence of the C5 basin (\ph$\approx$-160\degree,\ps$\approx$160\degree) and instead predicted a large basin that incorporated both the C7$_{eq}$ and C5 basins. 

By generations 6 and 11, the PMF contours closely resemble the ground truth, indicating a significant improvement in the accuracy of the model. For easier visualization of the errors, the difference heatmaps are shown below each PMF plot to illustrate the deviations from the ground truth, with blue indicating underestimation and red overestimation of the PMF values. As can be seen from the figure, there is a marked reduction in the deviation to the ground truth PMF from Gen~1 to Gen~6, which is also reflected in the mean absolute error (MAE) values in Fig.~\ref{fig:pmfs}b. This is then followed by a plateau from Gen~6 (or Gen~3) onwards, suggesting that the models have learned the phase space well early in the active learning iterations. 
Albeit only small improvements, we still observe slight refinement of PMF contours on the bottom left quadrant of the deviation heatmap beyond Gen~6.

\subsection{Ensemble-based Uncertainty}

Beyond the GMM-based uncertainty, we also explored the implementation of our workflow utilizing ensemble-derived uncertainty as the CV, as shown in Sec.~\ref{sec:ensemble}.
As the ensemble uncertainty (standard deviation) is directly related to the true error, no CP was used. 
However, as discussed in Sec.~\ref{sec:ensemble_eabf-gamd}, the simulations are very sensitive to the choice of the coupling constant $\sigma$. 
Specifically, tight couplings lead to early terminations of the simulations, whereas soft biasing, has almost no guiding effect. 
%Therefore, much more conservative settings have to be used with the ensemble-based uncertainty. 
This effect arises from the ensemble uncertainty's non-linear behavior, which displays strong confidence within known regions while extrapolative predictions rapidly diverge, leading to pronounced uncertainty fluctuations.
In contrast, the GMM NLL measure is smoother by design.
Consequently, after 10 active learning iterations, the extent of conformational space exploration under ensemble uncertainty was significantly less than that achieved using GMM uncertainty, as demonstrated in the previous section.

\subsection{Comparison with Uncertainty as Biasing Potential}

Prior to this work (see Sec.~\ref{sec:RelatedWork}), uncertainty has already been used to direct sampling by lowering the energy of configurations proportional to their predicted uncertainty (Eq.~\ref{eq:AttractiveBias}).
% Hence, we call that an attractive biasing method.
This method is straightforward to implement without needing any enhanced sampling algorithm, however, this is also one of its largest drawbacks.
During the first iterations, the uncertainty of most configurations is high, lowering all energy barriers and almost pulling the system away from known regions. 
In later iterations, however, the uncertainty close to the metastable regions is low, and thus does not alter the PES close to them significantly, which can confine the simulations in these states leading to no further exploration.
Contrarily, most enhanced sampling methods are able to push the system along the uncertainty gradient towards high uncertainty regions.

To validate this, we conducted active learning employing Eq.~\ref{eq:AttractiveBias} with GMM-based uncertainty (see Sec.~\ref{sec:BiasingEnergyNoCV}), instead of utilizing eABF-GaMD.
The initial iterations, as depicted in Fig.~\ref{fig:attractivebias_al}, yield a substantial number of new and informative configurations. However, the rate of discovering new configurations markedly decreases after the first three cycles, with minimal additional data acquired in the subsequent five iterations, as opposed to GMM-uncertainty eABF-GaMD that continuously increase exploration even after Gen~7 (Fig.~\ref{fig:perf}).
This results in important regions of the \ph-\ps plane remaining unexplored.
Furthermore, the force testing error remains higher compared to uncertainty-guided exploration and the energy error on the test set exhibit considerable fluctuations.

%%%%%%%%%%%%%%%%%%%
%%% CONCLUSION %%%%
%%%%%%%%%%%%%%%%%%%
\section{Discussion and Conclusion}

In this study, we introduce a new perspective on curating training sets for MLIPs by incorporating uncertainty as a collective variable (CV) for enhanced sampling methods. 
This approach harnesses the predictive uncertainty inherent in NNIPs as a novel CV, thereby facilitating the exploration of conformational space. 
This approach has three key advantages: 
First, it enables an exploration of configuration space independently of predefined human-crafted CVs. 
Second, the approach overcomes the restrictions of limited degrees of freedom typically associated with most conventional CVs. This is because the predicted uncertainty from an NNIP can represent the global state of the system, thereby offering a more expansive and adaptable representation of achievable configurations. 
Third, utilization of uncertainty allows for a dynamic adjustment of the sampling focus. By incorporating uncertainty as a CV, our method utilizes uncertainty as a gauge to direct simulations towards regions of the potential energy landscape that are less well-defined and more critical for the NNIP. These could be extrapolative regions or transition states regions, which can be different from each other. 
Our proposed active learning protocol, which employs single-model uncertainty and thus eliminates need for ensembles, has been shown to outperform prior approaches in both effectiveness of potential energy surface exploration and computational efficiency.

% Future refinements could include stringent thresholds for selecting new configurations, particularly when dealing with ground truth calculations that are more computationally expensive. Additionally, periodically increasing the dimensionality of the MLIP's embedding could ensure it remains sufficiently expressive to accommodate the increasingly diverse data.

%The uncertainty-guided enhanced sampling method is shown on the alanine dipeptide molecule. In the alanine dipeptide molecule, the approach was able to explore the phase space well and achieve a coverage fraction comparable to a 1200~K unbiased simulation within 10 active learning iterations and maximum enhanced sampling temperature of 500~K.

% \section*{Acknowledgements}

% \section{Impact Statement}
% % does not count towards the page limit
% % See https://icml.cc/Conferences/2024/CallForPapers and change as needed
% This work presents a novel active learning approach to collect highly informative molecular configurations when constructing interactomic potentials for atomistic simulations.
% In light of the technical nature of our work, we feel there are no specific ethical or societal consequences that warrant special attention.

\bibliography{references}
\bibliographystyle{icml2024}

%%%%%%%%%%%%%%%%%%%%%%%%%%%%%%%%%%%%%%%%%%%%%%%%%%%%%%%%%%%%%%%%%%%%%%%%%%%%%%%
%%%%%%%%%%%%%%%%%%%%%%%%%%%%%%%%%%%%%%%%%%%%%%%%%%%%%%%%%%%%%%%%%%%%%%%%%%%%%%%
% APPENDIX
%%%%%%%%%%%%%%%%%%%%%%%%%%%%%%%%%%%%%%%%%%%%%%%%%%%%%%%%%%%%%%%%%%%%%%%%%%%%%%%
%%%%%%%%%%%%%%%%%%%%%%%%%%%%%%%%%%%%%%%%%%%%%%%%%%%%%%%%%%%%%%%%%%%%%%%%%%%%%%%
\newpage
\appendix
\onecolumn

\renewcommand{\thefigure}{S\arabic{figure}}
\renewcommand{\thetable}{S\arabic{table}}
\renewcommand{\theequation}{S\arabic{equation}}
\setcounter{figure}{0}

\section{Methods}
\subsection{MACE architecture} \label{sec:mace}
All neural network interatomic potentials (NNIPs) were trained using the MACE architecture, which at its core has an equivariant message-passing scheme that incorporates higher body order messages\cite{batatia_mace_2023}. 
The MACE code was obtained from the public package (\url{https://github.com/ACEsuit/mace.git}), version 0.3.0. 
In this work, the MACE models used to run uncertainty-guided enhanced sampling are made up of two layers, 4-channel dimension for tensor decomposition and $L = 1$ maximal message equivariance, with a spherical harmonics expansion up to $l_{max} = 3$ and a body-order correlation of 3. 
For model training, the AMSGrad variant of the Adam optimizer was used with an initial learning rate of 0.01, and the \texttt{ReduceLROnPlateau} scheduler with a patience of 30 and a decay factor of 0.8. 
In addition, we applied the effect of stochastic weight averaging (SWA) after 800 epochs with a learning rate of 0.001 and the exponential moving average (EMA) effect with a decay factor of 0.99. 
All other hyperparameters use the defaults specified in the public package. 
Note that the MACE models used for production runs and generating the potential of mean force (PMF) use a channel dimension of 16 instead of 4-channels. 
All other hyperparameters used are the same. 

\subsection{Details of eABF-GaMD sampling}\label{sec:eabf-gamd}

The Adaptive Biasing Force (ABF) method is a common technique used in molecular dynamics simulations to overcome barriers in free energy landscapes. 
This method involves applying a continuously updated biasing force along the CV of interest, $\xi(\mathbf{x})$ that cancels the running estimate of average force, which, over time, completely flattens the effective potential of mean force (PMF) along the CV\cite{comer_adaptive_2015}. 
Here, we seek to replace the CV of interest with the predicted uncertainty, $\xi(\mathbf{x}) = u(\mathbf{x})$. 
The biasing force is taken to be the opposite of the running estimate of the force along the CV
\begin{align}
    -\langle \mathbf{F}_{\xi} \rangle_{z} &= \frac{1}{N_\text{sampl}(\xi=z)} \sum_{i=1}^{N_\text{sampl}(\xi=z)} \frac{\partial V(\mathbf{x}_i)}{\partial \xi} \nonumber \\
     &= - \frac{1}{N_\text{sampl}(\xi=z)} \sum_{i=1}^{N_\text{sampl}(\xi=z)} \mathbf{F}_i \cdot \frac{\nabla \xi(\mathbf{x}_i)}{|\nabla \xi(\mathbf{x}_i)|^2}
\end{align} 
where $V(\mathbf{x})$ denotes the PES, $N_\text{sampl}(\xi=z)$ the number of samples collected, for which the CV has the value $z$, and $\mathbf{F}_i$ the Cartesian forces for configuration $\mathbf{x}_i$.

The GaMD and eABF combined PES reads
\begin{align}
        V_{\text{ext}}(\mathbf{x}, \lambda, t) &= V(\mathbf{x}) + \frac{1}{2\beta\sigma^2}(u(\mathbf{x}) - \lambda)^2 
        + V_\text{ABF}(\lambda, t) + V_{GaMD}(\mathbf{x})
\end{align}
where $V_{ABF}(\mathbf{x})$ is the ABF potential and $V_{GaMD}(\mathbf{x})$ is the GaMD potential defined by
\begin{equation}
    V_{GaMD} (\mathbf{x}) = \begin{cases}
        \frac{1}{2} k (E - V(\mathbf{x}))^2 &, V(\mathbf{x}) < E \\
        0 &, V(\mathbf{x}) \geq E\ .
        \end{cases}
\end{equation}
The energy threshold is set to the upper bound value, $E = V_\text{min} + (1/k)$, and the harmonic force constant, $k = \left(1 - (\sigma_0/\sigma_V)\right) ((V_\text{max} - V_\text{min})/(V_\text{avg} - V_\text{min})$ following Ref \cite{miao_gaussian_2015}. 
$V_\text{max}$, $V_\text{min}$, $V_\text{avg}$, and $\sigma_V$ are the maximum, minimum, average, and standard deviation of the potential energy of the unbiased system, respectively. 
$\sigma_0$ was set to 0.01. 

All uncertainty-guided eABF-GaMD simulations were performed with a time step of 0.25~fs. 
The Langevin thermostat with a friction coefficient of 1.0~ps$^{-1}$ was used in all simulations to maintain the temperature.
Starting velocities were drawn randomly from the respective Maxwell-Boltzmann distribution.
All simulations were automatically stopped if the minimum distance between all atoms was lower than 0.75~{\AA} or when distances between bonded atoms (Fig \ref{fig:fig1}a) exceeded 2.0~{\AA}. 

%%%%%%%%%%%%
% Data set %
%%%%%%%%%%%%
\subsection{Data sets} \label{sec:datasets}
The training and test sets were generated with the Amber ff19SB force field\cite{tian_ff19sb_2020} using the OpenMM software\cite{eastman_openmm_2017}. 
We conducted two NVT simulations of alanine dipeptide in vacuum at 500~K and 1200~K. 
Both production runs had a length of 1~ns, samples were collected every 0.2~ps, resulting in 5000 frames from each simulation. 
For the training set, 100 configurations were selected at random from the 500~K simulations. 
The remaining 9900 configurations, encompassing those from both the 500~K and 1200~K simulations, were randomly split into the calibration (100 frames) and test set (9800 frames).

\subsection{Umbrella sampling and MBAR protocols} \label{sec:umbrella}
In order to gauge NNIP accuracy beyond testing error and mere simulation stability, we compare the PMF along the dihedrals \ph and \ps in alanine dipeptide obtained from the ground truth with those simulated with the 16-channel models.
We chose umbrella sampling\cite{torrie_nonphysical_1977} for this task such that we could compute partial PMFs in case NNIP simulations are unstable (especially likely for early generations).

For the \enquote{ground truth} PMF, the conformational space of \ph and \ps was subdivided into 48 windows, each 7.5$^\circ$ apart, resulting in a total of 2304 simulations. 
Each simulation ran for 105~ps at 300~K, with the initial 5~ps trajectory discarded as equilibration steps, and dihedral angles of the configurations were recorded every 0.025~ps. 
The time step was 1~fs, no bonds were frozen.
These simulations were generated  using the python interface of the OpenMM software\cite{eastman_openmm_2017} together with the Amber ff19SB force field\cite{tian_ff19sb_2020}. 
In each simulation, a harmonic restraint was applied centered at the current \ph-\ps window with a force constant 146~kJ mol$^{-1}$ rad$^{-2}$.

For NNIP derived PMFs, each of the \ph and \ps dihedral was divided into 30 windows, yielding a total of 900 simulations. 
After 1.25~ps of equilibration, we performed 2.5~ps of production run in each window.
Frames were collected every 10 steps with a single time step of 0.25~fs. 
The force constant for the harmonic restraint on \ph-\ps was 192~kJ mol$^{-1}$ rad$^{-2}$. 
These NNIPs-based simulations were carried out using the Atomistic Simulation Environment (ASE) package\cite{hjorth_larsen_atomic_2017}. 

The unbiased Boltzmann weights were computed with the Multistate Bennett Acceptance Ratio (MBAR) method\cite{kumar_weighted_1992,shirts_statistically_2008,shirts_statistically_2020}.
We used the MBAR code from the Adaptive-Sampling package available at https://github.com/ochsenfeld-lab/adaptive\_sampling.git\cite{hulm_statistically_2022} to aggregate data, obtain Boltzmann weights for all frames, and construct the PMFs. 
The convergence criterion for the MBAR method was set to $10^{-6}$.

\section{Uncertainty-guided enhanced sampling using ensembles-based uncertainty} \label{sec:ensemble}
As a baseline comparison, we also included experiments of uncertainty-guided eABF-GaMD using ensemble-based uncertainty. In the ensemble-based models, uncertainty is assumed to derive from various sources of randomness in the training of several independent NNs, which include parameter initializations, optimizations, or hyperparameter choices\cite{heid_characterizing_2023}. For an ensemble that consists of $M$ distinct NNs, the predicted energy, $\mu_E(\mathbf{x})$ and forces, $\pmb{\mu}_F(\mathbf{x})$ of each configuration, $\mathbf{x}$ are taken as the arithmetic mean of the predictions from each individual NN
\begin{align}
    \mu_E(\mathbf{x}) &= \frac{1}{M} \sum^M_m \hat{E}_m(\mathbf{x}), \\
    \pmb{\mu}_F(\mathbf{x}) &= \frac{1}{M} \sum^M_m \hat{\mathbf{F}}_m(\mathbf{x}),
\end{align}
The uncertainty estimates can then be taken from the standard deviation of predictions from each individual NNs in an ensemble\cite{lakshminarayanan_simple_2017}, given by
\begin{align}
    \sigma_E(\mathbf{x}) &= \sqrt{\frac{1}{M-1} \sum^M_m (\hat{E}_m(\mathbf{x}) - \mu_E(\mathbf{x}))^2} \\
    \sigma_{F_i}(\mathbf{x}) &= \sqrt{\frac{1}{M-1} \sum^M_m \left[ \frac{1}{3} \Vert \hat{\mathbf{F}}_{i}(\mathbf{x}) - \pmb{\mu}_{F_{i}}(\mathbf{x}) \Vert^2_2 \right]}\label{eq:ens_unc}\\
    \sigma_F(\mathbf{x}) &= \frac{1}{n} \sum_i^{n} \sigma_{F_i}(\mathbf{x})
\end{align}
where $n$ is the number of atoms in each configuration, $\sigma_E(\mathbf{x})$ is the energy standard deviation, $\sigma_{F_i}^2(\mathbf{x})$ is the standard deviation of force for atom $i$, and $\sigma_F^2(\mathbf{x})$ is the standard deviation for the entire configuration.

\subsection{Workflow details} \label{sec:ensemble_details}
In this section, we employ an ensemble consisting of five neural networks (NNs), each based on the MACE architecture but with varying hyperparameters as follows:

\begin{itemize}
    \item 2 layers, 4-channel dimension with a body correlation order of 3
    \item 3 layers, 4-channel dimension with a body correlation order of 2
    \item 2 layers, 5-channel dimension with a body correlation order of 2
    \item 3 layers, 4-channel dimension with a body correlation order of 3
    \item 2 layers, 5-channel dimension with a body correlation order of 2
\end{itemize}
Except for these specified differences, all other hyperparameters remain consistent with those outlined in Sec.~\ref{sec:mace}. It is important to note that each NN in the ensemble is trained and validated on different splits of the training set, and we have intentionally chosen \enquote{smaller} models to facilitate faster exploration in the enhanced sampling simulations.

For the uncertainty-guided eABF-GaMD simulations, we adopt a \enquote{thermal coupling width}, $\sigma$ distinct from the value used in GMM-based uncertainty active learning. This variation and its implications are further explored in Sec.~\ref{sec:ensemble_eabf-gamd}. We have also set $u_\text{cutoff} = 0.25$ in the first active learning iteration, and incrementally decreased by a factor of 0.9 after every iteration. Note that this is also markedly different from the GMM-based uncertainty method, a distinction that will be elaborated upon in Sec.~\ref{sec:ensemble_eabf-gamd}. All remaining parameters for the ensemble based enhanced sampling are the same with those used in the GMM-based uncertainty framework, including two eABF-GaMD simulations at 300~K and 500~K at each active learning iteration. No conformal prediction is done on ensemble-based uncertainty since the variance is representative of the true error.

\subsection{Analysis of eABF-GaMD for ensemble-based uncertainty} \label{sec:ensemble_eabf-gamd}
\begin{figure}[htbp]
    \centering
    \includegraphics[width=\linewidth]{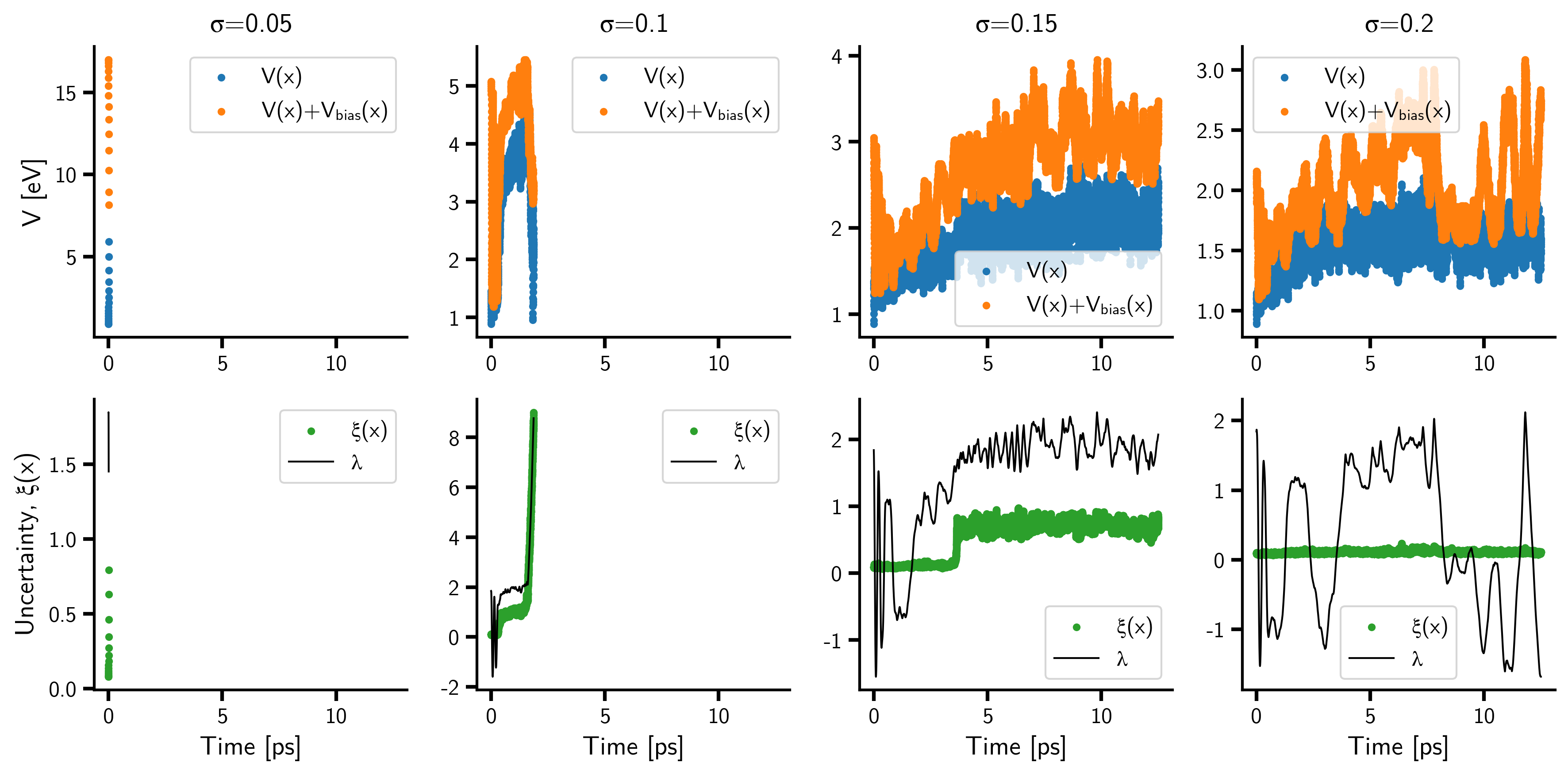}
    \caption{
    Exploration of the \enquote{thermal coupling width}, $\sigma$ parameter for guiding eABF-GaMD exploration at 300~K using ensemble-based uncertainty. The top row displays the predicted potential energy, $V(\mathbf{x})$ alongside the combined bias potential from eABF-GaMD, $V(\mathbf{x}) + V_\text{bias}(\mathbf{x})$ across simulation time. Plots in bottom row show the ensemble-predicted uncertainty, $\xi(\mathbf{x})$ and the fictitious variable, $\lambda$, plotted against simulation time. 
    }
    \label{fig:ensemble_eabf_params}
\end{figure}

In this section, we investigate the variation of the \enquote{thermal coupling width}, $\sigma$ for ensemble-based uncertainty method. 
We have observed in general that the ensemble-based uncertainty behaves very differently from the GMM-based uncertainty. 
As illustrated in the lower panels of Fig.~\ref{sec:ensemble_eabf-gamd}, the ensemble-based uncertainty is generally very low ($\sim0.01~$eV/\AA), and has a tendency to stay around the same values. 
This high confidence can be attributed to the ensemble models' lower prediction errors and their enhanced ability to generalize across configurations similar to those within the training set. 
However, once an extrapolative configuration is encountered, each NN in the ensemble yields diverging predictions, possibly due to different trends, gradients and curvatures in these extrapolative regions. 
This discrepancy then creates a steep \enquote{step} function of uncertainty as observed in the bottom plots, leading to unphysical structures that terminate the simulations — a phenomenon not observed with single-model approaches like GMM, which tend to produce smoother transitions across distorted configurations.

Consequently, our investigation into $\sigma$ values for ensemble-based uncertainty spans a narrower range (0.05 to 0.2) compared to the 0.5 value employed in GMM-based uncertainty (refer to Fig.~\ref{fig:ensemble_eabf_params}). 
The NN ensemble used for the simulations in this figure is trained on the initial training set that contains 100 configurations. 
At $\sigma=0.2$, the 12.5~ps simulation was completed successfully, but the resultant bias potential was insufficient to drive exploration into new regions, as confirmed by the flat uncertainty values, $\xi(x)$ in the corresponding plot. 
In the case of $\sigma = 0.05$ and $\sigma=0.1$, the bias potentials applied are able to guide the exploration towards extrapolative regime but led immediately to distorted configurations. 
A balance between exploration and simulation stability was achieved at $\sigma = 0.15$, which was subsequently adopted for the initial active learning iteration and reduced by a factor of 0.9 in each subsequent iterations.

The uncertainty threshold, $u_\text{cutoff}$ for data acquisition for ensemble-based uncertainty differs significantly from the value used for GMM-based uncertainty (where $u_\text{cutoff} = 2.0$). This is because we observed that the ensemble-based uncertainty estimate tends to stay consistently low during the eABF-GaMD simulations, which then necessitates distinct thresholds. Furthermore, $u_\text{cutoff}$ is decreased by a factor of 0.9 after every active learning iteration, unlike for GMM-uncertainty where $u_\text{cutoff}$ is increased by a factor of 1.05. This adjustment reflects the observation that, with successive iterations and an expanding dataset, the ensemble's predicted error in extrapolative regions decreases, rendering it increasingly challenging to induce bias even with tighter thermal coupling width. Further insights and detailed discussions on these findings will be presented in the subsequent section.

\subsection{Active learning}
\begin{figure}[htbp]
    \centering
    \includegraphics[width=\linewidth]{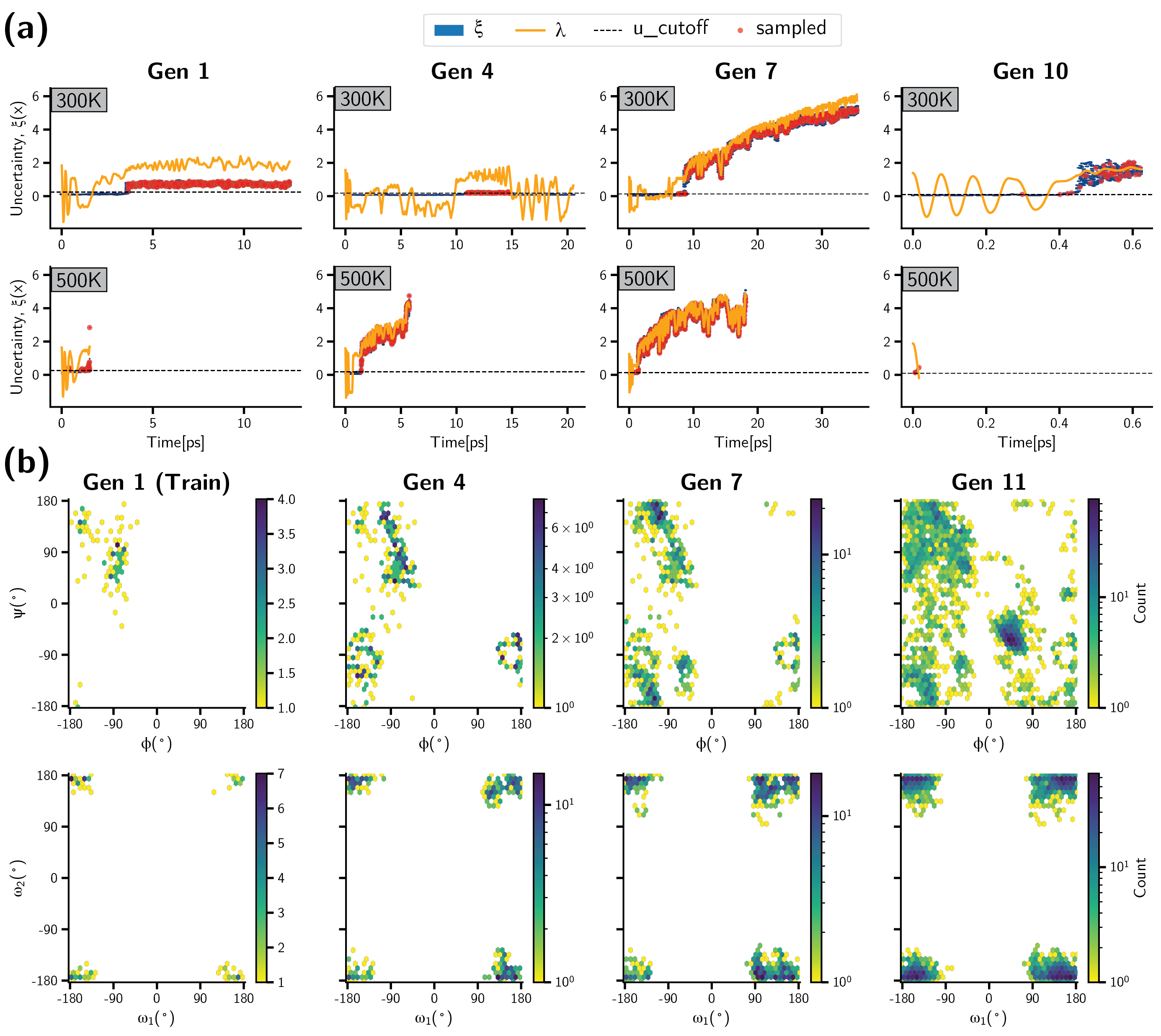}
    \caption{
    \textbf{(a)}, Ensemble-predicted uncertainty ($\xi$), the fictitious variable ($\lambda$), the uncertainty threshold (u\_cutoff), and the sampled points to be included for training of NNIPs in the next generation, plotted against the eABF-GaMD simulation time for different generations of the active learning iteration. 
    \textbf{(b)}, Hexbin plots representing data sets accumulative over generations of active learning, showcasing the distribution of \ph-\ps (top) and \omone-\omtwo (bottom) backbone dihedral angles during ensemble uncertainty-guided eABF-GaMD simulations. 
    }
    \label{fig:ensemble_al}
\end{figure}

Fig.~\ref{fig:ensemble_al} illustrates the outcomes of an active learning experiment using ensemble-based uncertainty. Observations from the simulations at 300~K for Gen~1 and 4 in Fig.~\ref{fig:ensemble_al}a and \ref{fig:ensemble_al}b reveal that the ensemble-derived uncertainty, $\xi$, tend to cluster around similar regions, despite the application of tight coupling, i.e., strong biasing. 
Conversely, in the 500~K simulations for these generations, the biasing potentials escalate excessively, leading to rapid formation of unphysical configurations that halt the simulations. With increasing active learning iterations, the extrapolative capabilities of the ensemble NNIPs show improvement and demonstrated enhanced exploration of the phase space (Fig.~\ref{fig:ensemble_al}b). This also sustained longer eABF-GaMD simulations. Despite these improvements, the expansion of explored regions was found to progress at a slower rate compared to the GMM-based uncertainty approach. Achieving optimal simulation stability across successive generations necessitates a more meticulous adjustment of the eABF-GaMD parameters.

\begin{figure}[htbp]
    \centering
    \includegraphics[width=\linewidth]{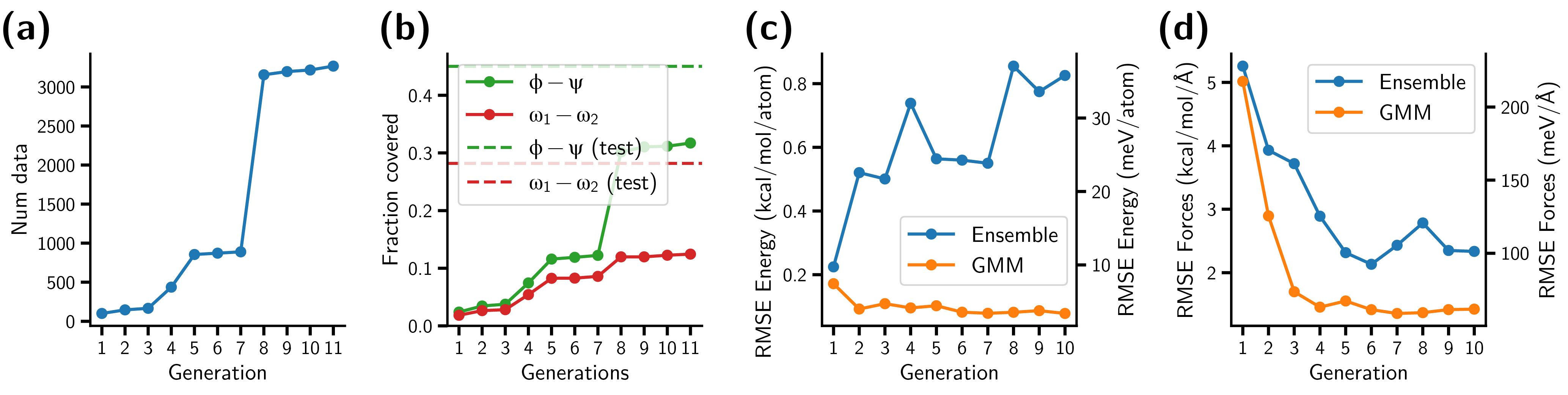}
    \caption{
    \textbf{(a)}, Number of total data points used for training the NNIPs in each generation. The initial value of 100 points at generation 1 indicates number of initial training data shown in Fig \ref{fig:fig1}a.
    \textbf{(b)}, Fraction of grid-based coverage of the \ph-\ps and \omone-\omtwo space in each generation. Horizontal dashed lines describes coverage for the test set. 
    \textbf{(c)} and \textbf{(d)}, panels depict the mean absolute error (MAE) in the predicted energy and forces for test set configurations over successive generations. Blue markers represent results from NN ensembles derived through active learning iterations utilizing an ensemble-based uncertainty approach. In contrast, orange markers denote outcomes from single NNs (4-channel dimension models) within the GMM-based uncertainty framework. It is important to note that, with the exception of the first generation, the NN ensembles and single NNs from the GMM uncertainty framework are trained on distinct datasets, each curated from their respective uncertainty-guided eABF-GaMD simulations. The dual axes in each plot provide measurements in the two standard units commonly employed in atomistic simulations with MLIPs.
    }
    \label{fig:ensemble_perf}
\end{figure}

To have a more qualitative understanding of the coverage of phase space, the number of data sampled for training and the \ph-\ps and \omone-\omtwo coverage fractions are shown in Fig.~\ref{fig:ensemble_perf}a and b. As observed, the expansion in coverage fraction and volume of sampled data is modest up until Gen~7, with exploration predominantly concentrated in areas of low uncertainty. A notable surge in exploration occurs in Gen~7, attributed to an optimal thermal coupling constant that not only facilitated exploration of new regions but also maintained simulation stability, as was also detailed in Fig.~\ref{fig:ensemble_al}a (third column). Despite conducting 10 active learning cycles, the overall coverage of phase space remains limited. 

In terms of accuracy on the test set (Figs.~\ref{fig:ensemble_perf}c and \ref{fig:ensemble_perf}d), the NN ensembles exhibit gradual improvement in force predictions but deteriorate in energy estimations. Additionally, the NN ensembles consistently underperform compared to single NNs GMM-based uncertainty approach, likely due to a less diverse training dataset. Notably, from Gen~7 onwards, despite the training data encompassing a broader section of the phase space, both energy and force prediction errors increase, indicating a complexity in the relationship between phase space coverage and predictive accuracy.

\section{Employing Uncertainty as Biasing Energy Rather than Collective Variable}
\label{sec:BiasingEnergyNoCV}

To evaluate the effectiveness of our uncertainty-guided eABF-GaMD approach, we conducted a comparative analysis with methodologies presented in the works of \citeauthor{kulichenko_uncertainty-driven_2023}, \citeauthor{van_der_oord_hyperactive_2023}, and \citeauthor{zaverkin_uncertainty-biased_2023}, which similarly employ uncertainty for phase space exploration. In these studies, the authors propose the implementation of a bias energy, $V_\text{bias}(\mathbf{x})$, which is directly correlated with the uncertainty of a given configuration, $\mathbf{x}$. Each of these works adopts a distinct function to formulate $V_\text{bias}(\mathbf{x})$, reflecting slight variations in their approach to integrate uncertainty into the biasing mechanism. For simplification, we adopted the following strategy to bias the system
\begin{align}
    \tilde{V}(\mathbf{x}) &= V(\mathbf{x}) + V_\text{bias}(\mathbf{x}) \nonumber \\ 
    &= V(\mathbf{x}) - \gamma u(\mathbf{x})
    \label{eq:AttractiveBias}
\end{align}
where $V(\mathbf{x})$ is the potential energy of the configuration as provided by the MLIP, $\gamma$ is the biasing strength, and $u(\mathbf{x})$ is the predicted uncertainty of the corresponding configuration. 

An active learning experiment using uncertainty as biasing energy would then follow an analogous procedure compared to the eABF-GaMD approach of training a NNIP, conducting simulations biased by uncertainty, acquire data points from these simulations, performing ground truth calculations on the newly gathered data points, and finally retraining the NNIPs with the updated dataset. In this case, no eABF-GaMD simulations are required. 
The NN architecture used for this study has the same hyperparameter specifications as outlined in Sec.~\ref{sec:mace}, and the uncertainty is computed from a GMM as CP-adjusted NLL as in the main body of this work unlike in \citeauthor{kulichenko_uncertainty-driven_2023} and \citeauthor{van_der_oord_hyperactive_2023}, where ensemble-based uncertainty is used. Additionally, we have optimized the biasing strength, $\gamma$, to effectively bias the exploration towards novel regions without yielding nonviable configurations, as discussed in the next section.

\subsection{Biasing strength optimization}

\begin{figure}[htbp]
    \centering
    \includegraphics[width=\linewidth]{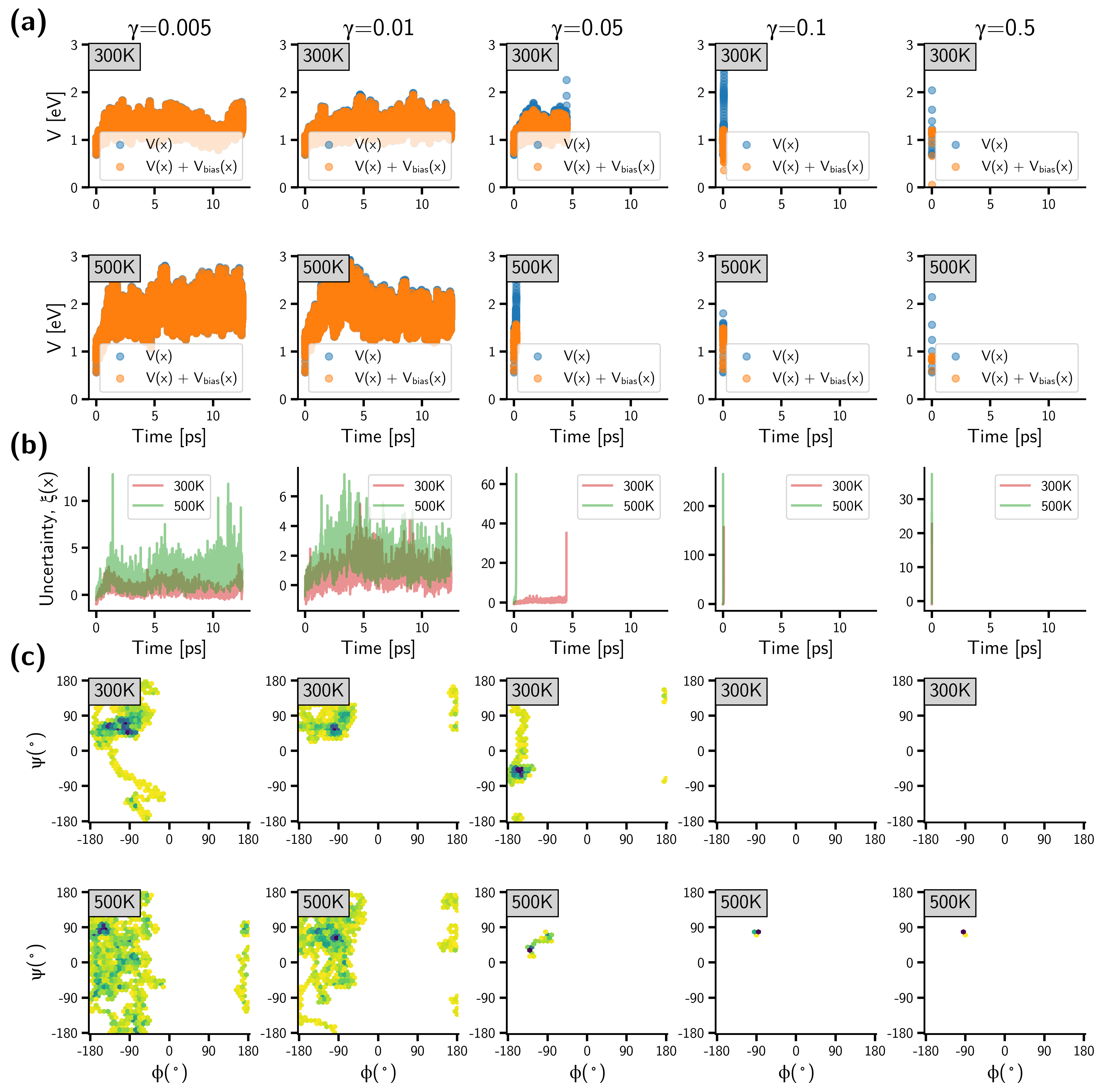}
    \caption{
    Investigation of the biasing strength, denoted by the parameter $\gamma$, ranging from 0.005 (leftmost) to 0.5 (rightmost), to effectively steer exploration towards extrapolative regions.
    \textbf{(a)}, The potential energy, $V(\mathbf{x})$ and combined bias potential, $V(\mathbf{x}) + V_\text{bias}(\mathbf{x})$ plotted against simulation time for simulations at 300~K (top row) and 500~K (bottom row). 
    \textbf{(b)}, Evolution of GMM-based uncertainty, $\xi(x)$ throughout simulations conducted at 300~K (upper row) and 500~K (lower row). 
    \textbf{(c)}, Hexbin plots representing \ph-\ps backbone dihedral angles of configurations simulated at 300~K and 500~K. Color bars are removed due to space constraints.
    }
    \label{fig:attractivebias_params}
\end{figure}

To optimize the balance between exploration and exploitation, we analyzed the impact of biasing strength on simulations conducted at temperatures of 300K and 500K, as depicted in Fig.~\ref{fig:attractivebias_params}. For biasing strengths $\gamma > 0.1$, the bias potentials were excessively high, leading to the premature termination of simulations due to the immediate generation of highly distorted, and thus highly uncertain, configurations. Conversely, at $\gamma < 0.05$, the biasing forces were moderate enough to guide the exploration towards novel regions while avoiding the creation of unphysical configurations. Notably, a biasing strength of $\gamma = 0.005$ resulted in more effective exploration than $\gamma = 0.01$, with higher predicted uncertainties observed despite similar magnitudes of biasing potentials. We speculate that a lower biasing strength enables a more incremental exploration of the potential energy surface, preventing the simulation from bypassing barriers and recurrently settling into identical energy minima.

For a more comparative analysis, we also performed active learning experiments using this uncertainty-based biasing energy approach. We adopted $\gamma = 0.005$ based on the findings of the optimization study, and increased the biasing strength by a factor of 1.3 with each increasing generation. The uncertainty threshold, $u_\text{cutoff}$ is also decreased by a factor of 0.9 after each iteration, with the initial $u_\text{cutoff}$ set to 2.0. The active learning experiment is repeated 5 times, and the highest coverage fraction experiment is shown in the subsequent section.

\subsection{Active learning}

\begin{figure}[htbp]
    \centering
    \includegraphics[width=\linewidth]{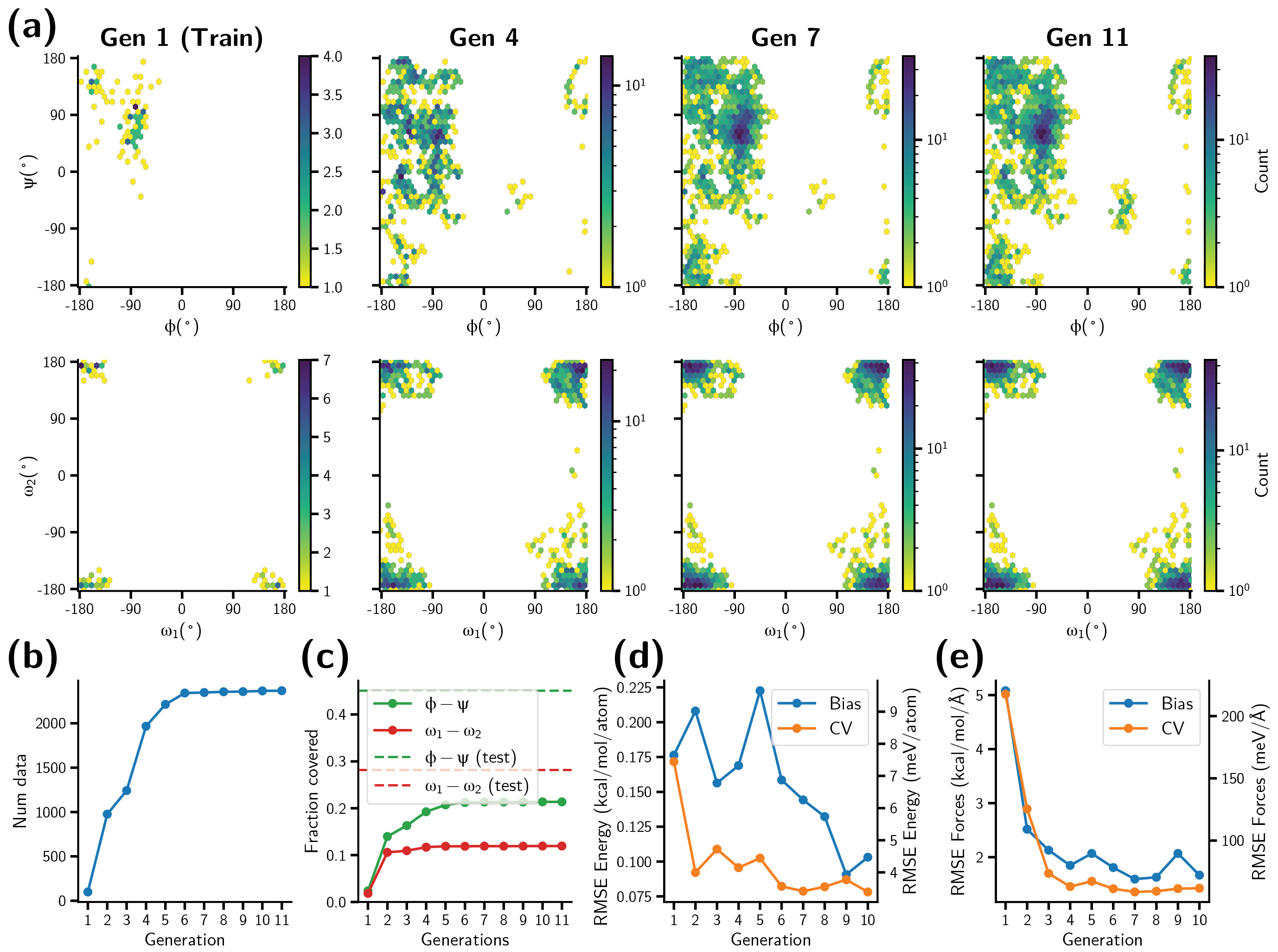}
    \caption{
    \textbf{(a)}, Hexbin plots representing data sets accumulative over generations of active learning, showcasing the distribution of \ph-\ps (top) and \omone-\omtwo (bottom) backbone dihedral angles during uncertainty as biasing energy simulations.
    \textbf{(b)}, Number of total data points used for training the NNIPs in each generation. The initial value of 100 points at generation 1 indicates number of initial training data shown in Fig \ref{fig:fig1}a.
    \textbf{(c)}, Fraction of grid-based coverage of the \ph-\ps and \omone-\omtwo space in each generation. Horizontal dashed lines describes coverage for the test set. 
    \textbf{(d)} and \textbf{(e)}, panels depict the mean absolute error (MAE) in the predicted energy and forces for test set configurations over successive generations. Blue markers represent results from NNIPs derived through active learning iterations utilizing uncertainty as biasing energy approach (Bias), whereas orange markers denote outcomes from NNIPs with the GMM-based uncertainty framework (CV). It is important to note that, with the exception of the first generation, the two NNIPs are trained on distinct datasets, each curated from their respective simulations. The dual axes in each plot provide measurements in the two standard units commonly employed in atomistic simulations with MLIPs.
    }
    \label{fig:attractivebias_al}
\end{figure}

In the active learning iterations using uncertainty as a biasing energy, we can observe from Fig.~\ref{fig:attractivebias_al}a that the exploration for $\phi < 0^\circ$ is relatively extensive by Gen~4, whereas areas with $\phi > 0^\circ$ remains largely unexplored. Note that the explored regions are distinct from those depicted in Fig.~\ref{fig:attractivebias_params}c since the starting configurations and exploration paths in every active learning experiment are random. At generations beyond Gen~4, we see that the coverage fraction appears to stagnate, despite a systematic reduction in the uncertainty threshold $u_\text{cutoff}$ for data acquisition with every iteration. Nonetheless, the prediction errors shown in Fig.~\ref{fig:attractivebias_al}d and \ref{fig:attractivebias_al}e continuously decrease with generations, suggesting that the data points acquired in successive iterations are effectively enriching the NNIPs' extrapolative capabilities in the potential energy surface. Another interesting observation is the limited capability of the simulations to explore the non-planar \omone and \omtwo backbone dihedral angles, which is also observed in the ensemble-based uncertainty for eABF-GaMD simulations. 

\section{Additional figures}

\begin{figure}[htbp]
    \centering
    \includegraphics[width=\linewidth]{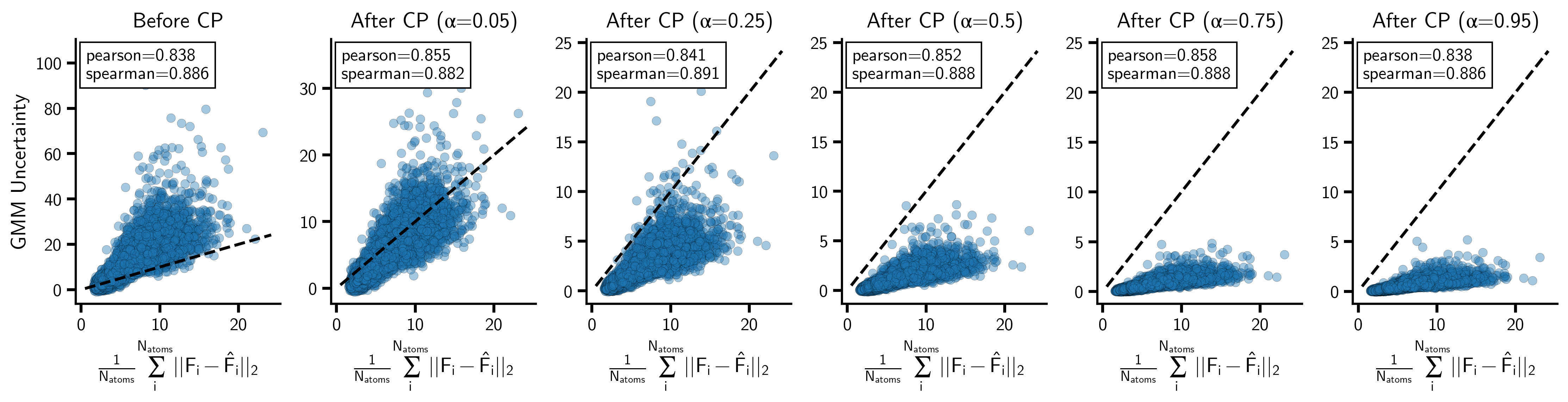}
    \caption{
    Conformal prediction on the GMM-based uncertainty values using different $\alpha$ values, calibrated to the true error of NNIP's predictions, $\frac{1}{N_\text{atoms}} \sum_i^{N_\text{atoms}} ||\mathbf{F}_{i} - \mathbf{\hat{F}}_{i}||_2$ on the calibration set (See Sec.~\ref{sec:cp} for more details). Values in the text boxes on upper left corner of plots show the Pearson and Spearman correlation coefficients. 
    }
    \label{fig:cp_alpha}
\end{figure}

\begin{figure}
    \centering
    \includegraphics[width=\linewidth]{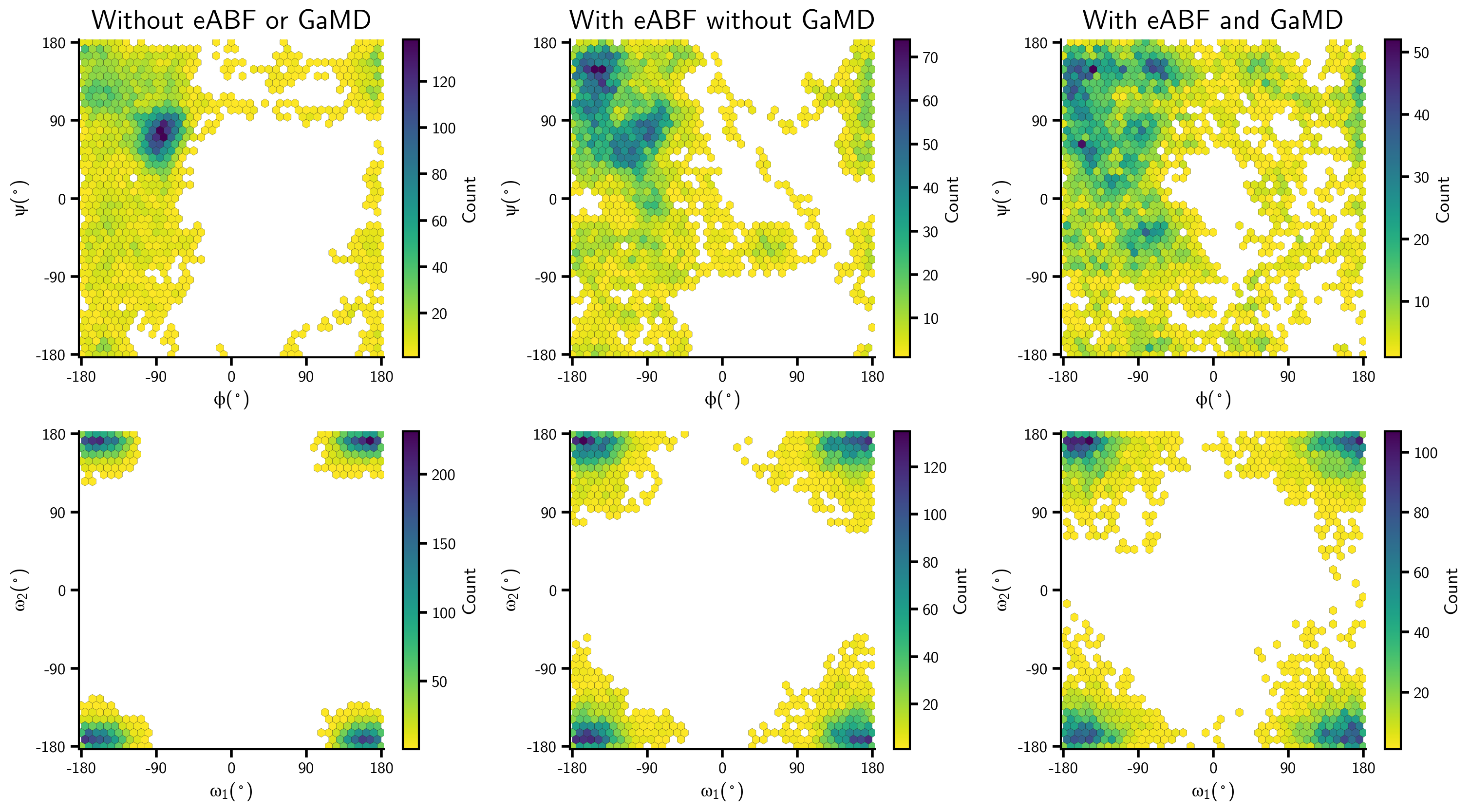}
    \caption{Coverage of explored configurations in the \ph-\ps (top row) and \omone-\omtwo (bottom row) space during 10 simulations each of unbiased NVT (leftmost: without eABF or GaMD), with uncertainty-guided eABF (middle), and with uncertainty-guided eABF-GaMD (rightmost) at 500~K. All simulations completed successfully (10~ps) without forming unphysical configurations.}
    \label{fig:comparison_eabf-GaMD-500K}
\end{figure}

%%%%%%%%%CONFORMAL PREDICTION%%%%%%%%%%%%%%%%%

%%%%%%%%%%%%%%%%%%%%%%%%%%%%%%%%%%%%%%%%%%%%%%%%%%%%%%%%%%%%%%%%%%%%%%%%%%%%%%%
%%%%%%%%%%%%%%%%%%%%%%%%%%%%%%%%%%%%%%%%%%%%%%%%%%%%%%%%%%%%%%%%%%%%%%%%%%%%%%%

\end{document}